\documentclass[sigconf]{acmart}
\AtBeginDocument{%
  }

\setcopyright{cc}
\setcctype{by}
\copyrightyear{2026}
\acmYear{2026}
\acmDOI{10.1145/3770855.3818104}
\acmISBN{979-8-4007-2258-5/2026/08}
\acmConference[KDD '26]{Proceedings of the 32nd ACM SIGKDD Conference
  on Knowledge Discovery and Data Mining V.2}{August 09--13,
  2026}{Jeju Island, Republic of Korea}
\acmBooktitle{Proceedings of the 32nd ACM SIGKDD Conference on Knowledge
  Discovery and Data Mining V.2 (KDD '26), August 09--13, 2026, Jeju Island, Republic of Korea}

\newcommand{\sstitle}[1]{\noindent\textbf{#1.\/}}

\def\Snospace~{\S{}}

\usepackage{multirow}
\usepackage{booktabs}
\usepackage{algorithm}
\usepackage{algorithmicx}
\usepackage{graphicx}\graphicspath{{figures/}}
\usepackage{algpseudocode}
\usepackage{amsmath}
\usepackage{float}
\usepackage[shortlabels]{enumitem}
\usepackage{subcaption}
\usepackage[skip=1pt]{caption}
\makeatletter
\newcommand{\removelatexerror}{\let\@latex@error\@gobble}
\makeatother

\begin{document}

%\title{An Evaluation of Multiple Models for Unlabeled Text-to-SQL Data}
%\title{Cost-Effective and Label-Free Evaluation of LLM-based Text2SQL models}
%\title{MetaSQLEvaluator: Evaluating Text2SQL Performance on Unseen Databases Without Labels}
% \title{Cost-Effective Evaluation of Multiple Text2SQL Models on Unseen Databases}

% \title{A Cost-Effective Framework for Model Evaluation on Unlabeled Data using Meta-Learning}

\title{Learning to Evaluate: Cost-Effective Model Evaluation on Unlabeled Data with Meta-Learning}

% \affiliation{%

\author{Trinh Pham}
\email{phkhanhtrinh23@gmail.com}
\affiliation{%
  \institution{Griffith University}
  \city{Gold Coast}
  \country{Australia}}

\author{Viet Huynh}
\email{v.huynh@ecu.edu.au}
\affiliation{%
  \institution{Edith Cowan University}
  \city{Perth}
  \country{Australia}}

\author{Hongzhi Yin}
\authornote{Hongzhi Yin and Quoc Viet Hung Nguyen are co-corresponding authors.}
\email{h.yin1@uq.edu.au}
\affiliation{%
  \institution{The University of Queensland}
  \city{Brisbane}
  \country{Australia}}

\author{Quoc Viet Hung Nguyen}
\authornotemark[1]
\email{quocviethung1@gmail.com}
\affiliation{%
  \institution{Griffith University}
  \city{Gold Coast}
  \country{Australia}}

\author{Thanh Tam Nguyen}
\email{thanhtamlhp@gmail.com}
\affiliation{%
  \institution{Griffith University}
  \city{Gold Coast}
  \country{Australia}}

\renewcommand{\shortauthors}{Trinh Pham, Viet Huynh, Hongzhi Yin, Quoc Viet Hung Nguyen, and Thanh Tam Nguyen}

\begin{abstract}
The rapid advancement of machine learning has led to an unprecedented expansion of model ecosystems, making it increasingly difficult to assess the reliability of newly released models on unseen and unlabeled data. Existing evaluation pipelines typically rely on costly annotation, repeated fine-tuning, or assumptions that do not generalize well to new models. We introduce \textit{MetaEvaluator}, a cost-effective, model-agnostic framework for fast, label-free evaluation of unseen models across diverse architectures and modalities. \textit{MetaEvaluator} meta-learns over a pool of reference models to acquire an effective initialization for accurate assessment of unseen models, thereby amortizing evaluation cost and eliminating the need for per-model retraining. To the best of our knowledge, this is the first model-agnostic framework that evaluates new models on unlabeled datasets. Extensive experiments demonstrate that \textit{MetaEvaluator} delivers stable and accurate performance estimates at substantially lower cost than conventional approaches, enabling scalable benchmarking on unlabeled datasets for emerging models. The code is available at: \url{https://github.com/phkhanhtrinh23/MetaEvaluator}.

\end{abstract}






\maketitle

\section{Introduction}
\label{sec:intro}

Recent progress in machine learning is driven by large pretrained model families and rapidly growing data collections, most of which remain unlabeled~\cite{liu2025unlabeled,he2025reevaluating}. This trend introduces a core deployment problem for organizations: choosing among many newly released and unseen models for an unlabeled workload. Recent systems now span multilingual transfer, paraphrasing, and increasingly capable Text2SQL agents~\cite{le2024lampat,pham2024unibridge,pham2025multilingual,pham2026avsql}, further accelerating model turnover in practice. Consider an organization that deploys a new Text2SQL model to query an internal enterprise database. No labeled question--SQL pairs exist, and manual annotation would require domain experts and weeks of effort, making rapid model selection impractical. This setting exposes a double challenge: the model is unseen, and the target dataset is entirely unlabeled. We specifically target deployments where labeling the target workload may be infeasible, e.g., when expert annotation is prohibitively costly, when data grow or change too rapidly for a label-centric method to adapt, or when privacy constraints restrict access to data for annotation. In such regimes, constructing a labeled evaluation for each model is impractical, motivating a label-free estimator. Most existing evaluation pipelines nevertheless operate on one model at a time. Some require human or pseudo-labeling~\cite{boyeau2025autoeval, angelopoulos2023prediction, fisch2024stratified}. Others rely on repeated fine-tuning~\cite{Schelter2020, yu2022predicting, jiang2022assessing, chen2021detecting, deng2021labels}. Several approaches work for only one model~\cite{deng2021labels,guillory2021predicting,zheng2023gnnevaluator}.
\begin{figure}[t]
    \centering
    \includegraphics[width=0.85\linewidth]{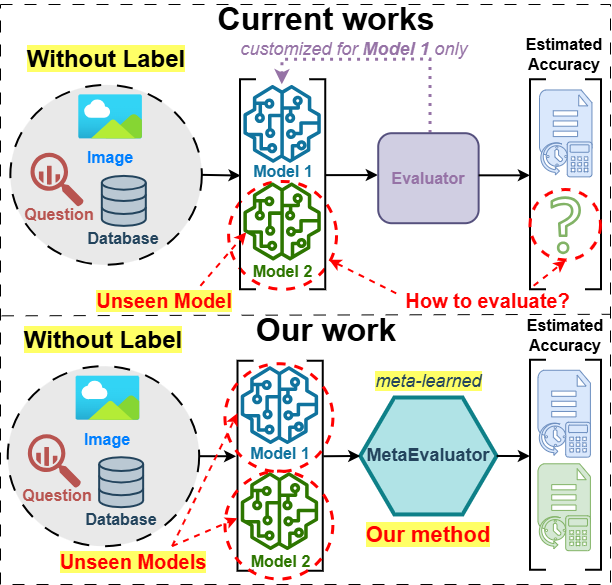}
    \caption{Unlike existing methods that struggle to assess unseen models without labels, MetaEvaluator leverages meta-learning to efficiently estimate accuracy for unseen models.}
    \label{fig:motivation}
    \vspace{-1em}
\end{figure}
Judge-based systems using large language models (LLMs) have also been proposed~\cite{gu2024survey, zheng2023judging, Liu2025nl2sqlbugs}. However, these techniques are designed for fixed architectures and model-specific behaviors, and they incur substantial computational cost~\cite{salinas2025tuning} and human labor overhead~\cite{YangVLDB2018,wang-etal-2021-want-reduce} when applied to each new model. Inspired by findings~\cite{schurholt2022model, zhang2023model} that pretrained systems exhibit structured and predictable performance trends across architectures and domains rather than arbitrary variation, we pose the central question of this work: \textit{Can we learn to evaluate unseen models on unlabeled data by transferring knowledge from previously evaluated models?}

As illustrated in \autoref{fig:motivation}, we answer this question by introducing \textbf{MetaEvaluator}, a model-agnostic framework designed to generalize performance estimation to newly arriving, unseen models on unlabeled target workloads without relying on extensive human annotation or repeated per-model training. MetaEvaluator reframes evaluation as a meta-learning problem: it learns transferable performance patterns from a shared pool of reference models that have been systematically evaluated across diverse datasets, architectures, and distribution shifts. By distilling these patterns into compact context representations, MetaEvaluator can rapidly adapt to a new model on an unlabeled dataset. Across all settings, MetaEvaluator produces predictions that closely track ground-truth performance while substantially reducing evaluation overhead. This design amortizes cost across reference models and enables scalable assessment in rapidly evolving model ecosystems. Our main contributions are:
\begin{itemize}
    % \item \textit{Formulation:} We formulate the problem of label-free performance estimation for unseen models under distribution shift across heterogeneous architectures and modalities.
    % \item \textit{Method:} We introduce \textbf{MetaEvaluator}, a model-agnostic meta-learning framework that amortizes evaluation cost by learning transferable performance patterns from a shared pool of reference models, enabling rapid and reliable deployment decisions for newly arriving systems.
    % \item \textit{Dataset:} We develop \textbf{MetaDataset}, a large-scale dataset that captures diverse and systematic distribution shifts and supports controlled evaluation across model families and learning domains.
    \item \textit{Formulation:} As model architectures and datasets evolve rapidly, we formulate the double challenge of evaluating unseen models on unseen and unlabeled data. We design a method that generalizes across heterogeneous architectures and modalities, including Text2SQL and Image Classification.
    \item \textit{Method:} We propose \textbf{MetaEvaluator}, a model-agnostic meta-learning framework that learns how performance varies across models and shifts by transferring knowledge from a pool of reference models, enabling rapid adaptation to newly released architectures on unlabeled workloads without per-model retraining.
    \item \textit{Dataset:} We introduce \textbf{MetaDataset}, a large-scale and systematically constructed corpus of model--shift pairs that spans Text2SQL and Image Classification, providing diverse unlabeled deployment scenarios and accurate performance supervision for MetaEvaluator.
    % \item \textit{Method:} We address this problem with \textbf{MetaEvaluator}, a model-agnostic meta-learning framework that learns transferable performance patterns from a shared pool of reference models, supervised by \textbf{MetaDataset}--a large-scale dataset designed to capture diverse and realistic unlabeled distribution shifts.
    \item \textit{Benchmarking:} \textbf{MetaEvaluator} enables a lightweight, fast, and automated benchmarking framework that can ingest newly released models and promptly return accurate performance estimates on unlabeled workloads, supporting rapid deployment cycles and model-selection feedback in real organizational settings.
    % This framework amortizes training cost across reference models and eliminates the need for costly annotation or per-model retraining during deployment.
    
    % \item \textcolor{blue}{ Lightweight framework to benchmark new models.} %We design MetaEvaluator to be lightweight and cost-effective without per-model training or costly annotations. We report latency to support comparisons with existing evaluation pipelines.
\end{itemize}

\section{Related Work}
\label{sec:related}

Most existing label-free evaluation methods assess a single fixed model and do not handle the harder setting in which both the model and the target dataset are unseen at deployment time. Model ecosystems expand rapidly and pipelines that rely on per-model fine-tuning quickly become impractical. To the best of our knowledge, no prior work addresses this \textit{double challenge}: evaluating unseen models on unlabeled data across modalities. Therefore, we study both Text2SQL and Image Classification to pursue a unified solution that remains effective as architectures evolve.

% In Image Classification, early work such as AutoEval~\cite{deng2021labels} trains a regression-based evaluator that maps distribution distances between training data and synthetically shifted variants in a model's representation space to accuracy estimates. DoC~\cite{guillory2021predicting} replaces distributional distances with differences in confidence scores, aiming to improve robustness under natural distribution shifts. While effective for a fixed backbone and unlabeled test sets, both approaches are customized to the evaluated model, requiring retraining of the evaluator when a new architecture is introduced. SelfTrainEns~\cite{chen2021detecting} estimates accuracy through agreement patterns among auxiliary ensembles trained on the same task, but this substantially increases computational cost because multiple models must be trained solely for evaluation. ATC~\cite{garg2022leveraging} further reduces overhead by selecting a confidence threshold on labeled validation data and transferring it to unlabeled workloads. However, the threshold is still model-specific and must be re-tuned for every new system, and for Text2SQL-based autoregressive models, aggregating token-level confidences can introduce additional evaluation latency.

In Image Classification, AutoEval~\cite{deng2021labels} and DoC~\cite{guillory2021predicting} estimate accuracy from representation-level distribution distances or confidence shifts, but both are trained for a specific backbone and must be retrained for each new architecture. SelfTrainEns~\cite{chen2021detecting} estimates accuracy through agreement patterns among auxiliary ensembles trained on the same task, but this substantially increases computational cost because multiple models must be trained solely for evaluation. ATC~\cite{garg2022leveraging} further reduces overhead by selecting a confidence threshold and transferring it to unlabeled workloads. However, the threshold is still model-specific and must be re-tuned for every new system. A separate line of work, including AGD~\cite{jiang2022assessing} and PseudoAutoEval~\cite{boyeau2025autoeval}, relies on retraining the target model or generating pseudo-labels, which introduces substantial computational overhead, additional inference passes, and human annotation.

However, these approaches are largely developed for Image Classification. In Text2SQL, where schemas and query distributions evolve rapidly with scarce labels~\cite{Li2024dawn,pham2025multilingual,pham2026avsql,pham2026efficient}, label-free evaluation of unseen models remains unexplored. NL2SQL-BUGS~\cite{Liu2025nl2sqlbugs} targets fine-grained debugging with both automated detectors and human-in-the-loop judgment, and therefore supports only low-throughput analysis at the query level and requiring substantial human involvement.

\sstitle{Meta-Learning}
Meta-learning, or \textit{learning to learn}, optimizes a model across a distribution of tasks so that it can adapt to a new task from only a few examples. Optimization-based methods are the most relevant to our setting. MAML~\cite{finn2017maml} learns an initialization from which a few gradient steps yield strong task-specific performance, but its bi-level objective requires differentiating through the inner-loop updates, which is costly and unstable as the adapted parameter set grows. FO-MAML~\cite{finn2017maml} and Reptile~\cite{nichol2018reptile} drop or approximate these second-order terms to reduce cost, while Meta-SGD~\cite{li2017metasgd} additionally meta-learns per-parameter learning rates. ANIL~\cite{raghu2020anil} observes that most of MAML's benefit comes from feature reuse and adapts only the task-specific head, freezing the shared body during the inner loop. Metric-based methods such as ProtoNet~\cite{snell2017prototypical} instead avoid inner-loop optimization by classifying against learned class prototypes. We build on this optimization-based view but treat each \textit{reference model} (rather than each dataset) as a task, and adapt only a compact model-specific context vector while keeping the evaluator backbone shared (\autoref{sec:metaevaluator}).

Overall, as summarized in \autoref{fig:motivation}, prior methods are model-specific and costly, preventing scalable evaluation under the double challenge of unseen models and unlabeled data. We close this gap with a meta-learning framework that learns evaluators across reference models and shifts, enabling rapid, model-agnostic performance estimation for both Image Classification and Text2SQL.

\section{Problem Formulation}
\label{sec:problem_formulation}

\begin{figure*}[t]
    \centering
    \includegraphics[width=0.65\linewidth]{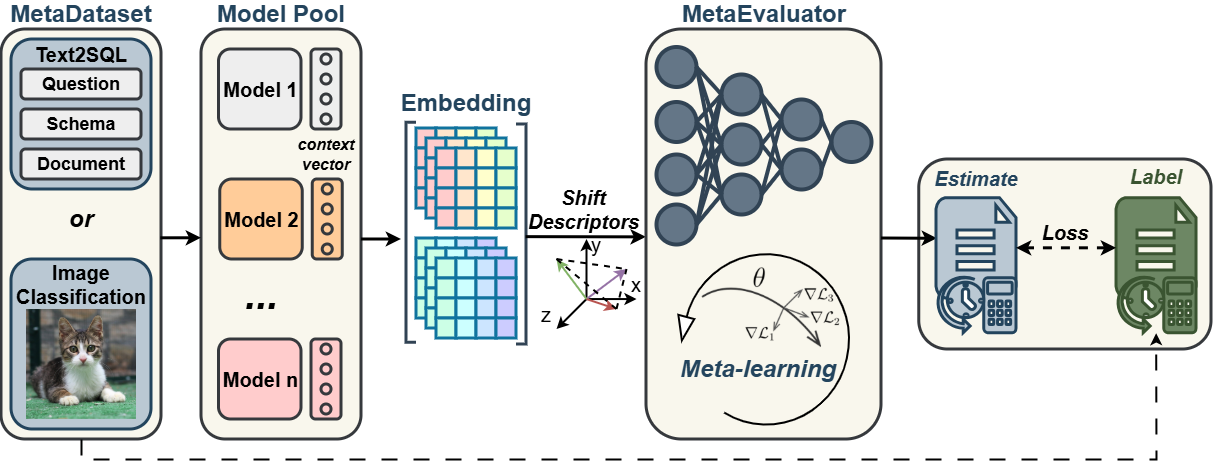}
    \caption{MetaEvaluator applies meta-learning over a pool of reference models, using data from MetaDataset to learn how to map shift descriptors to estimates by adapting to each reference model and minimizing error against known performance.}
    \label{fig:training_pipeline}
    \vspace{-1em}
\end{figure*}

% \textcolor{blue}{Formulate from machine learning, to meta-learning, to our method.}
Motivated by the limitations of prior work, we formalize a setting in which both the evaluated model and the target dataset are unseen and unlabeled at deployment time. To address this double challenge, we start from classical supervised learning and extend it to label-free evaluation and meta-learning. In the single-model setting, we train a supervised evaluator as a regressor over dataset-level signals that summarize the model's behavior on the target workload: train--test mismatch and the model's true accuracy. We then use meta-learning by aggregating such supervised evaluation problems across many reference models so that the evaluator can adapt to new architectures on unlabeled workloads.
Throughout this work, $n$ denotes the number of samples in a single dataset or workload, whereas $N$ denotes the number of dataset-level evaluation pairs used to train the evaluator.

\subsection{Model}
We begin with standard supervised learning, where one is given a labeled dataset $\mathcal{D}=\{(x_i,y_i)\}_{i=1}^n$ and a predictive model $f_\theta$ with parameters $\theta$, and training minimizes the empirical risk $\sum_{i=1}^n \mathcal{L}(f_\theta(x_i),y_i)$, where $\mathcal{L}$ denotes a loss such as mean-squared error (MSE). At evaluation time, the model is tested on a labeled set $\mathcal{D}_{\mathrm{test}}=\{(x_j^{\mathrm{test}},y_j^{\mathrm{test}})\}_{j=1}^{n'}$, producing predictions $\hat{y}_j=f_\theta(x_j^{\mathrm{test}})$, and robustness to distribution shift is quantified through an aggregate error such as mean absolute error $\mathrm{MAE}=\frac{1}{n'}\sum_{j=1}^{n'} |f_\theta(x_j^{\mathrm{test}})-y_j^{\mathrm{test}}|$.

% We begin with standard supervised learning. Given a labeled dataset $\mathcal{D}=\{(x_i,y_i)\}_{i=1}^n$ and a predictive model $f_\theta$ parameterized by $\theta$, training minimizes an empirical risk $\sum_{i=1}^n \mathcal{L}(f_\theta(x_i),y_i)$, where $\mathcal{L}$ denotes the loss function, e.g., mean-squared error (MSE). At evaluation time, the model is tested on $\mathcal{D}_{\mathrm{test}}=\{(x_j^{\mathrm{test}},y_j^{\mathrm{test}})\}_{j=1}^{n'}$, producing predictions $\hat{y}_j=f_\theta(x_j^{\mathrm{test}})$. Robustness to distribution shift is quantified through an aggregate error, such as mean absolute error $\mathrm{MAE}=\frac{1}{n'}\sum_{j=1}^{n'} |f_\theta(x_j^{\mathrm{test}})-y_j^{\mathrm{test}}|$, computed across diverse test distributions.

\sstitle{Label-free Model Evaluation}
We now turn from learning predictive models to learning evaluators. For a trained model $f(\cdot\mid\psi)$ with training set $\mathcal{D}_S$ and a labeled evaluation workload $\mathcal{D}_T$, prior works~\cite{deng2021labels,guillory2021predicting} construct a shift descriptor $\mathrm{SD}$ summarizing distributional or confidence-level shifts between $\mathcal{D}_S$ and $\mathcal{D}_T$, and pair it with the model's true accuracy $a$. This yields a training set:
\begin{equation}
\mathcal{D}_{\mathrm{train}}=\{(\mathrm{SD}_j,a_j)\}_{j=1}^N,
\label{eq:eval_dataset}
\end{equation}
from which an evaluator $g_\theta$ is trained by minimizing an empirical regression loss $\sum_{j=1}^N \mathcal{L}(g_\theta(\mathrm{SD}_j), a_j\big)$. At deployment, for a new unlabeled target workload $\mathcal{D}_{T}^{\mathrm{new}}$, one computes $\mathrm{SD}_{\mathrm{new}}$ between $\mathcal{D}_S$ and $\mathcal{D}_{T}^{\mathrm{new}}$, predicts $\hat{a}=g_\theta(\mathrm{SD}_{\mathrm{new}})$, and later measures generalization by comparing $\hat{a}$ with the unknown true accuracy $a_{\mathrm{new}}$ once labels become available offline.

\sstitle{Scaling to Unseen Models}
When evaluation must extend to a set of unseen models $\{m_k\}$, a naive strategy repeatedly acquires labeled target workloads $\mathcal{D}_T^{(m_k)}$ and forms the evaluator training set with new pairs $(\mathrm{SD}^{(m_k)},a^{(m_k)})$, followed by supervised fine-tuning of $g_\theta$. This procedure amounts to optimizing a single parameter vector $\theta^{\star}=\arg\min_\theta \sum_k \sum_j \mathcal{L}(g_\theta(\mathrm{SD}_j^{(m_k)}), a_j^{(m_k)}\big)$, which aggregates losses across all previously observed models. Consequently, predictions for a novel architecture $m_{\mathrm{new}}$ are forced through this global solution $\theta^{\star}$, effectively using its shift descriptors with an evaluator that lacks task-specific adaptation.

\sstitle{Meta-Learning for Model Evaluation}
To overcome this limitation, we cast evaluation itself as a meta-learning problem. We construct a reference pool of models $\mathcal{M}$ and treat each $m\in\mathcal{M}$ as a separate task. For every $m$, we form a task-specific dataset:
\begin{equation}
\mathcal{D}^{(m)}=\{(\mathrm{SD}_i^{(m)},a_i^{(m)})\}_{i=1}^{N_m},
\label{eq:task_dataset}
\end{equation}
where $\mathrm{SD}_i^{(m)}$ captures the shift between the training set of $m$ and a labeled workload drawn from \textit{MetaDataset} (later introduced in \autoref{sec:method}), and $a_i^{(m)}$ is the corresponding accuracy. The collection of all such tasks defines the meta-training set $\mathcal{S}=\{\mathcal{D}^{(m)}: m\in\mathcal{M}\}$. We meta-learn parameters $\theta$ for an evaluator $g_\theta$ so that, after a small number of adaptation steps on a new labeled workload $\mathcal{D}_{new}$, the adapted evaluator $g_{\theta_m}$ predicts the accuracy for that model $m$. In this way, $\theta$ is optimized not to evaluate any single architecture, but to encode a transferable strategy for evaluating unseen ones. At test time, we are given an unseen model, parameterized as $f_{\mathrm{new}}(\cdot\mid\psi_{\mathrm{new}})$, and an unlabeled target workload:
\begin{equation}
\mathcal{D}_T=\{x_i^T\}_{i=1}^n ,
\label{eq:target_dataset}
\end{equation}
on which the model produces predictions $\hat{y}_i=f_{\mathrm{new}}(x_i^T\mid\psi_{\mathrm{new}})$. Let $y_i^{T\star}$ denote the unknown ground-truth outputs and define the true dataset-level metric:
\begin{equation}
M^{\star}=\frac{1}{n}\sum_{i=1}^n m(\hat{y}_i,y_i^{T\star}),
\label{eq:true_metric}
\end{equation}
where $m(\cdot,\cdot)$ specifies a performance measure, e.g., exact match. Our goal is to estimate $M^{\star}$ without observing $y_i^{T\star}$. We compute shift descriptors between the training data of $f_{\mathrm{new}}$ and $\mathcal{D}_T$, adapt the meta-learned evaluator $g_\theta$ using a small set of reference signals if available, and output a prediction $\widehat{M}$. Performance is measured by the mean absolute error $|\widehat{M}-M^{\star}|$ across many unseen models and deployment shifts, reflecting whether the system has learned to evaluate rather than to memorize any specific architecture.

\subsection{Challenges}
Building on the formulation above, we study the double challenge of estimating the true performance $M^{\star}$ of an unseen model $f_{\mathrm{new}}(\cdot\mid\psi_{\mathrm{new}})$ on an unlabeled workload $\mathcal{D}_T$. This deployment scenario introduces several fundamental difficulties:
\begin{itemize}
  \item[$\xi1$]\textit{Absence of ground truth}: the labels $\{y_i^{T\star}\}$ for $\mathcal{D}_T$ are unavailable, so $M^{\star}$ in \autoref{eq:true_metric} cannot be computed directly.
  \item[$\xi2$]\textit{Cross-model generalization}: the evaluator $g_\theta$ must remain reliable when confronted with a novel model $f_{\mathrm{new}}(\cdot\mid\psi_{\mathrm{new}})$ whose behavior lies outside the reference pool $\mathcal{M}$ used during training.
  \item[$\xi3$]\textit{Distribution shift}: the target distribution $\mathcal{D}_T$ may differ substantially from the source data $\mathcal{D}_S$ in domain or input statistics, inducing unpredictable performance changes.
  \item[$\xi4$]\textit{Limited access}: the evaluator must operate without modifying the unseen model or its parameters $\psi_{\mathrm{new}}$.
  \item[$\xi5$]\textit{Efficiency constraints}: evaluation must remain lightweight in computation to enable scalable benchmarking and practical pre-deployment model selection.
\end{itemize}

\subsection{Objective}
Our objective is to estimate the true dataset-level performance $M^{\star}$ of an unseen model $f_{\mathrm{new}}(\cdot\mid\psi_{\mathrm{new}})$ on the unlabeled target workload $\mathcal{D}_T$, without access to the ground-truth labels $\{y_i^{T\star}\}$ or modifying the model itself, thereby directly addressing $\xi1$ and $\xi4$. The estimator must further satisfy efficiency constraints in deployment scenarios ($\xi5$).

Let $\mathrm{SD}_{\mathrm{src}}$ denote descriptors computed from $f_{\mathrm{new}}$ on its training set $\mathcal{D}_S$, and let $\mathrm{SD}_{\mathrm{tgt}}$ denote descriptors computed on the unlabeled target set $\mathcal{D}_T$. We summarize train--test differences through:
\begin{equation}
  \mathrm{SD}=h\big(\mathrm{SD}_{\mathrm{tgt}},\mathrm{SD}_{\mathrm{src}}\big),
  \label{eq:shift_descriptors}
\end{equation}
and seek an evaluator $g$, parameterized by $\theta$, that maps these descriptors to a performance estimate:
\begin{equation}
  \widehat{M}=g_\theta(\mathrm{SD}).
  \label{eq:evaluator}
\end{equation}

The evaluator is designed to achieve the following objectives:
\begin{enumerate}
  \item \textit{Accuracy} ($\xi1$--$\xi3$): minimize $|\widehat{M}-M^{\star}|$ across target workloads and unseen models.
  \item \textit{Uncertainty} ($\xi2$): provide calibrated uncertainty estimates, e.g., a prediction interval $[\widehat{M}-\delta_\alpha,\;\widehat{M}+\delta_\alpha]$ such that $\mathbb{P}(M^{\star} \in [\widehat{M}-\delta_\alpha,\;\widehat{M}+\delta_\alpha])\ge 1-\alpha$, where $\delta_\alpha$ is the interval half-width at miscoverage level $\alpha$.
  \item \textit{Generality} ($\xi1$, $\xi2$, $\xi4$): require no access to ground-truth labels and no changes to the unseen model or its parameters.
  \item \textit{Efficiency} ($\xi5$): operate with low runtime and resource consumption to enable scalable benchmarking.
\end{enumerate}

\section{Methodology}
\label{sec:method}

Motivated by Objectives~1--3 in \autoref{sec:problem_formulation}, we construct \textit{MetaDataset}, a unified multimodal corpus for meta-learning label-free evaluation. This exposes MetaEvaluator to varied shifts and model behaviors, enabling generalization to unseen architectures and supporting calibration through repeated observation across conditions. Our framework proceeds in two stages: (1) MetaDataset construction and (2) MetaEvaluator learning.

\subsection{MetaDataset}
\label{sec:metadataset}

Unlike conventional datasets that benchmark a fixed model on a single test distribution, MetaDataset is organized around model--shift pairs, providing multiple target workloads and true accuracies for each reference model and thus forming diverse meta-learning tasks. It must satisfy two general principles across modalities: (i) it should expose models to \textit{diverse and controllable distribution shifts} so evaluators can learn realistic performance degradation, and (ii) it should be \textit{scalable and low-barrier}, allowing practitioners to synthesize large volumes of shifted data without expert annotation. Practitioners are free to generate task-specific environments (e.g., graph, voice, medical image) and control the shifts. In this work, we instantiate this dataset for two representative domains: Text2SQL and Image Classification.

% \textcolor{blue}{Why we need MetaDataset and what's the difference between this and other datasets? Why is this useful to Meta-learning?}

% We construct MetaDataset, a unified corpus spanning Text2SQL and Image Classification. These tasks cover relational databases and natural images, enabling MetaEvaluator to learn across heterogeneous modalities and shifts.
% , and node classification 
% and graph-structured data,

\sstitle{Text2SQL}
We design relational environments that expose evaluators to schema evolution, SQL structural variation, and linguistic shift, reflecting deployment conditions in prior benchmarks.

\textit{Database diversity.} Starting from real-world tables (e.g., TabLib~\cite{eggert2023tablib} and KaggleDBQA~\cite{lee2021kaggledbqa}), we apply acquisition, refinement, and synthesis steps, using GPT-5 as the backbone model, to remove noise, cluster compatible schemas, infer foreign keys, standardize columns, and construct multi-table databases with realistic connectivity.

\textit{SQL diversity.} We generate queries ranging from simple projections to nested analytics. To overcome template bias, we augment large-scale generators with SQLForge~\cite{guo2025sqlforge}, PARSQL~\cite{dai2025parsql}, and semantics-preserving rewriting~\cite{cui2025llm}, producing diverse join patterns, subqueries, and dialectal variants (SQLite, PostgreSQL, Snowflake).

\textit{Question diversity.} Each SQL query is paired with multiple natural-language realizations (e.g., formal, colloquial, conversational, vague, etc.) inspired by SynSQL-2.5M~\cite{li2025omnisql}, SParC~\cite{yu2019sparc}, and CoSQL~\cite{yu2019cosql}. We further inject realistic noise (distractors and modifiers) observed in KaggleDBQA and BIRD~\cite{li2023bird}, while preserving the underlying SQL intent.

\sstitle{Image Classification}
We construct a multi-stage pipeline for label-preserving image generation under realistic distribution shift, combining vision--language guidance with diffusion models.

\textit{Dataset preparation.} For each class in several vision benchmarks (\autoref{sec:experiment_data_coverage}), we collect diverse seed images and apply CLIP-based filtering~\cite{radford2021clip} to remove ambiguous or low-alignment samples.

\textit{Semantic edit.} Following EvolveDirector~\cite{zhao2024evolvedirector}, a vision--language controller proposes edits that are executed by a text-conditioned diffusion model~\cite{rombach2022latentdiffusion}. We instantiate five families of shifts: illumination, material/surface properties, camera perturbations, background relocation, and contextual changes (e.g., weather, occlusion, motion blur). Shift severity is controlled by activating one family (\textit{mild}), two to three (\textit{moderate}), or at least three including material plus background or context (\textit{strong}).

\textit{Validation and filtering.} A second CLIP-based alignment check removes label drift, ensuring semantic consistency under shift.

\sstitle{Summary}
MetaDataset is a unified multimodal dataset for MetaEvaluator to learn distribution-aware, label-free performance estimation, label-free performance estimation from diverse model--shift pairs, thus generalizing across models and deployment conditions.

\subsection{MetaEvaluator}
\label{sec:metaevaluator}

%%%%% Algorithm: Meta-learning MetaEvaluator
\begin{algorithm}[t]
\caption{Meta-learning MetaEvaluator $g_{\theta}$ with $\mathcal{M}_{\mathrm{train}}$.}
\begin{algorithmic}[1]
\State \textbf{Input:} $\mathcal{D}_{\mathrm{train}}$, $\mathcal{D}_{\mathrm{val}}$, model pool $\mathcal{M}_{\mathrm{train}}$, initial parameters $\theta$, initial context vectors $\{ctx_m\}_{m\in\mathcal{M}_{\mathrm{train}}}$, learning rates $\alpha_{\mathrm{inner}},\alpha_{\mathrm{outer}}$, epochs $E$.
\For{$e=1$ to $E$}
\For{$m \in \mathcal{M}_{\mathrm{train}}$}
\State \textcolor{gray}{\textit{// Inner loop on target subsets drawn from $\mathcal{D}_{\mathrm{train}}$:}}
\State $\mathcal{L}_{\mathrm{train}}(m)\gets 0$.
\For{each sampled $(\mathrm{SD}_i,M_i^{\star})$ from $\mathcal{D}_{\mathrm{train}}$}
\State $\mathcal{L}_{\mathrm{train}}(m)\gets \mathcal{L}_{\mathrm{train}}(m)+\bigl(g_{\theta}(\mathrm{SD}_i,ctx_m)-M_i^{\star}\bigr)^2$.
\EndFor
\State \textcolor{gray}{\textit{// RMSE loss:}}
\State $\mathcal{L}_{\mathrm{train}}(m)\gets \sqrt{\mathcal{L}_{\mathrm{train}}(m)/b_{\mathrm{train}}}$.
\State \textcolor{gray}{\textit{// Update $ctx_m$ ($\theta$ fixed)}:}
\State $ctx_m \gets ctx_m-\alpha_{\mathrm{inner}}\nabla_{ctx_m}\mathcal{L}_{\mathrm{train}}(m)$.
\\
\State \textcolor{gray}{\textit{// Outer loop on target subsets drawn from $\mathcal{D}_{\mathrm{val}}$}}:
\State $\mathcal{L}_{\mathrm{val}}\gets 0$.
\For{each sampled $(\mathrm{SD}_j,M_j^{\star})$ from $\mathcal{D}_{\mathrm{val}}$}
\State $\mathcal{L}_{\mathrm{val}}\gets \mathcal{L}_{\mathrm{val}}+\bigl(g_{\theta}(\mathrm{SD}_j,ctx_{m})-M_j^{\star}\bigr)^2$.
\EndFor
\EndFor
\State $\mathcal{L}_{\mathrm{val}} \gets \sqrt{\mathcal{L}_{\mathrm{val}}/b_{\mathrm{val}}}$.
\State \textcolor{gray}{\textit{// Update $\theta$ (all $ctx_m$ fixed):}}
\State $\theta \gets \theta-\alpha_{\mathrm{outer}}\nabla_{\theta}\mathcal{L}_{\mathrm{val}}$.
\EndFor
\State \textbf{Output:} $\theta^{\star}=\theta$, $\{ctx_m^{\star}\}_{m\in\mathcal{M}_{\mathrm{train}}}$.
\end{algorithmic}
\label{alg:meta_learning_algorithm}
\end{algorithm}

\begin{algorithm}[t]
\caption{Evaluation for $m_{\mathrm{new}}$ with MetaEvaluator $g_{\theta^{\star}}$.}
\begin{algorithmic}[1]
\State \textbf{Input:} an unseen model $m_{\mathrm{new}}$, initial context $ctx_{\mathrm{new}}$, labeled split $\mathcal{D}_{\mathrm{train}}$, unseen and unlabeled data $\mathcal{D}_{T}$, optimal parameters $\theta^{\star}$, learning rate $\alpha$, adaptation steps $K$.
\For{$k$ =1 to $K$}
\State $\mathcal{L} \gets 0$.
\State \textcolor{gray}{\textit{// Inner loop on target subsets drawn from $\mathcal{D}_{\mathrm{train}}$:}}
\For{each sampled $(\mathrm{SD}_i,M_i^{\star})$ from $\mathcal{D}_{\mathrm{train}}$}
\State $\mathcal{L} \gets \mathcal{L}+\bigl(g_{\theta^{\star}}(\mathrm{SD}_i,ctx_{\mathrm{new}})-M_i^{\star}\bigr)^2$.
\EndFor
\State $\mathcal{L} \gets \sqrt{\mathcal{L}/b_{\mathrm{adapt}}}$.
\State \textcolor{gray}{\textit{// Update $ctx_{\mathrm{new}}$ ($\theta^{\star}$ fixed):}}
\State $ctx_{\mathrm{new}}\gets ctx_{\mathrm{new}}-\alpha \nabla_{ctx_{\mathrm{new}}}\mathcal{L}$.
\EndFor
\\
% \State \textcolor{gray}{\textit{// OT alignment of $ctx_{\mathrm{new}}$:}}
% \State $ctx_{\mathrm{OT}} \gets \textsc{AlignOT}(ctx_{\mathrm{new}})$.
\State \textcolor{gray}{\textit{// Compute $\mathrm{SD}^{unlb}$ and Estimate:}}
\State $\widehat{M}=g_{\theta^{\star}}(\mathrm{SD}^{unlb},ctx_{\mathrm{new}})$.
\State \textbf{Output:} $\widehat{M}$.
\end{algorithmic}
\label{alg:evaluation_algorithm}
\end{algorithm}

To satisfy the objectives in \autoref{sec:problem_formulation}, we meta-learn an evaluator over the broad coverage provided by MetaDataset (\autoref{sec:metadataset}) across diverse model architectures and distribution shifts.

\sstitle{Context Vector}
At the core of MetaEvaluator is a \textit{context vector} $ctx_m$: a lightweight, model-specific embedding that is passed as an additional \emph{input} to the global evaluator, i.e., predictions take the form $g_{\theta}(\mathrm{SD},ctx_m)$. The context vector encodes the task-specific identity of model $m$ (e.g., architecture family, training regime, and dataset characteristics), while the backbone parameters $\theta$ remain a single shared evaluation function across all models and are never specialized per model. This design departs from MAML~\cite{finn2017maml}, where every task adapts the full parameter vector $\theta_i$ in the inner loop, so updates for one model can interfere with those for another. By adapting only the low-dimensional $ctx_m$ in the inner loop, MetaEvaluator concentrates the gradient signal on a clean, model-specific subspace and avoids cross-task interference. Further results are reported in \autoref{sec:experiment_ablation}.

\sstitle{Meta-Learning}
As shown in \autoref{alg:meta_learning_algorithm}, we partition MetaDataset into two i.i.d.\ splits, $\mathcal{D}_{\mathrm{train}}$ and $\mathcal{D}_{\mathrm{val}}$, and optimize MetaEvaluator over a reference model pool $\mathcal{M}_{\mathrm{train}}$. Concretely, each episode corresponds to a single reference model $m$. During the inner loop, we randomly sample a batch of target subsets $\{\mathcal{D}_{T_i}\}$ from $\mathcal{D}_{\mathrm{train}}$. For each subset, we compute a shift descriptor $\mathrm{SD}(\mathcal{D}_S^{m},\mathcal{D}_{T_i})$ between the outputs of $m$ on its original training data $\mathcal{D}_S^{m}$ and on $\mathcal{D}_{T_i}$. Together with the corresponding true dataset-level performance $M_i^{\star}$, these sampled $(\mathrm{SD}, M_i^{\star})$ pairs form the meta-set for that episode and define the inner-loop loss used to adapt the model-specific context vector $ctx_m$ while keeping the global parameters $\theta$ fixed. During the outer loop, we analogously sample target subsets $\{\mathcal{D}_{T_j}\}$ from $\mathcal{D}_{\mathrm{val}}$ and construct the corresponding $(\mathrm{SD}, M_j^{\star})$ pairs to update $\theta$ while holding all $\{ctx_m\}_{m\in\mathcal{M}_{\mathrm{train}}}$ fixed. The shift descriptor is computed from the model's hidden-space outputs when available, or, for black-box models exposing only text outputs, from the embeddings of a separate frozen encoder (e.g., BERT). Therefore, the same formulation applies regardless of model access. In contrast to MAML~\cite{finn2017maml}, we treat each model as a task rather than each dataset. The overall procedure is depicted in \autoref{fig:training_pipeline}.

\sstitle{Evaluation}
During evaluation for an unseen model $m_{\mathrm{new}}$, we adapt only the newly initialized context vector $ctx_{\mathrm{new}}$ of $m_{\mathrm{new}}$, while keeping the globally meta-trained parameters $\theta^{\star}$ fixed. To do so, we reuse sampled target subsets $\{\mathcal{D}_{T_i}\}$ from the labeled split $\mathcal{D}_{\mathrm{train}}$ to compute the corresponding pairs $\{(\mathrm{SD}_i,M_i^{\star})\}$. We then perform a small number of lightweight adaptation steps on $ctx_{\mathrm{new}}$ by minimizing the resulting loss while holding $\theta^{\star}$ fixed, as detailed in \autoref{alg:evaluation_algorithm}. %We emphasize that the \emph{only} labeled dependency of the framework is this one-time adaptation of the context vector on target subsets drawn from the labeled split $\mathcal{D}_{\mathrm{train}}$ of MetaDataset; once $ctx_{\mathrm{new}}$ is fitted, estimating performance on the unseen and unlabeled target workload $\mathcal{D}_{T}$ requires no target labels and is therefore fully label-free.
Finally, given an unseen and unlabeled target workload $\mathcal{D}_{T}$, we compute $\mathrm{SD}^{unlb}$ as the distributional difference between the outputs of $m_{\mathrm{new}}$ on its original training data and $\mathcal{D}_{T}$, and estimate:
\begin{equation}
\widehat{M}=g_{\theta^{\star}}\!\left(\mathrm{SD}^{unlb},ctx_{\mathrm{new}}\right).
\label{eq:inference_prediction}
\end{equation}

\section{Experiment}
\label{sec:experiment}

Building a label-free evaluation method that remains reliable for unseen models under distribution shift raises four fundamental research questions:
\begin{itemize}
  \item \textbf{RQ1 (Data Coverage) (\autoref{sec:experiment_data_coverage}):} How well does MetaDataset capture diverse, realistic deployment scenarios and distribution shifts across tasks and domains?
  \item \textbf{RQ2 (Evaluator Learning) (\autoref{sec:experiment_evaluator_learning}):} How accurately does MetaEvaluator learn and generalize to unseen models and distribution shifts across modalities?
  \item \textbf{RQ3 (Benchmarking Capability) (\autoref{sec:experiment_benchmarking_capability}):} Does MetaEvaluator enable lightweight benchmarking as the number of unseen models and the reference model pool grow?
  \item \textbf{RQ4 (Component Utility) (\autoref{sec:experiment_ablation}):} How does each individual component of MetaEvaluator contribute to its overall estimation accuracy and robustness across unseen models and distribution-shifted workloads?
\end{itemize}

\sstitle{Environment} We run all experiments on a workstation with four NVIDIA GeForce RTX 4090 GPUs (24GB each) and an Intel Core i7-14700 CPU (20 cores, 2.1 GHz base). To maximize throughput during evaluator meta-learning, we use mixed-precision computation (\texttt{bfloat16}) and parallel data loading.

% \sstitle{Base Models}
% We evaluate MetaEvaluator on two learning domains: Text2SQL and Image Classification, using modality-specific pools of base models. In each domain, every model is fine-tuned on its training data until achieving its best validation performance. The resulting models are then applied to generate predictions and latent representations, which are aggregated into fixed-length vectors that summarize the train--test shift, called shift descriptors. These shift descriptors serve as inputs to MetaEvaluator for estimating performance on unlabeled target workloads.

\sstitle{Shift Descriptors}
We construct $\mathrm{SD}$ by concatenating 3 complementary hidden-space summaries: a Gaussian Fréchet term $\mathrm{SD}_F$, a Mahalanobis term $\mathrm{SD}_M$, and a sliced Wasserstein term $\mathrm{SD}_{SW}$, forming $\mathrm{SD}=[\mathrm{SD}_F,\mathrm{SD}_M,\mathrm{SD}_{SW}]$ as input to MetaEvaluator. $\mathrm{SD}_F$ captures global changes in embedding statistics, $\mathrm{SD}_M$ emphasizes rare or low-density examples such as uncommon SQL constructs or corrupted images, and $\mathrm{SD}_{SW}$ models directional geometric shifts caused by systematic changes in query structure or visual conditions. Together, these components provide a compact and expressive summary of train--test mismatch across modalities.

\sstitle{Training Settings}
MetaEvaluator is implemented as a three-layer MLP with hidden dimensions \{256, 128, 64\}, ReLU activations, and is trained to regress the true accuracy of a model from SDs. Training uses a batch size of 64 and the AdamW optimizer with learning rate $1{\times}10^{-4}$ (cosine decay), $\beta_1{=}0.9$, $\beta_2{=}0.999$, and weight decay $1{\times}10^{-3}$. MetaEvaluator performs meta-learning up to 100 epochs with early stopping based on validation MAE, applies dropout of 0.2 between hidden layers, and optimizes mean squared error (MSE) between predicted and true accuracies.

\sstitle{Evaluation Metrics}
We evaluate MetaEvaluator by measuring how accurately it predicts dataset-level performance on each target dataset. For Image Classification, the task metric is classification accuracy (Acc). For Text2SQL, we report both exact match (EM) and execution accuracy (EX). Given image data $\mathcal{D}=\{(x_i,y_i^\star)\}_{i=1}^{n}$: $\mathrm{Acc}(\mathcal{D}) = \frac{1}{n}\sum_{i=1}^{n}\mathbb{I}\!\left[\hat{y}_i=y_i^\star\right]$, where $\hat{y}_i$ is the prediction, $y_i^\star$ is the ground-truth label, and $\mathbb{I}[\cdot]$ is the indicator function. Given Text2SQL data $\mathcal{D}=\{(x_i,q_i^\star)\}_{i=1}^{n}$: $\mathrm{EM}(\mathcal{D}) = \frac{1}{n}\sum_{i=1}^{n}\mathbb{I}\!\left[\hat{q}_i=q_i^\star\right]$, where $\hat{q}_i$ is the predicted SQL, and $q_i^\star$ is the ground-truth SQL. Let $\mathrm{Exec}(\cdot)$ denote query execution on the database: $\mathrm{EX}(\mathcal{D}) = \frac{1}{n}\sum_{i=1}^{n}\mathbb{I}\!\left[\mathrm{Exec}(\hat{q}_i)=\mathrm{Exec}(q_i^\star)\right]$. Across $N$ dataset-level evaluation pairs, we quantify accuracy estimation quality using \textit{Mean Absolute Error} (MAE): $\mathrm{MAE} = \frac{1}{N}\sum_{i=1}^{N}\big|\widehat{M}_i-M_i^{\star}\big|$. We report MAE separately for Acc, EM, and EX.

\sstitle{Budget Constraints}
MetaDataset is generated under a fixed budget $B{=}1{,}000$ (USD). We decompose the total cost into three operations: (i) generation, (ii) filtering/validation, and (iii) execution (SQL only). For each modality $t\in\{\textsc{sql},\textsc{img}\}$, we partition the corpus into sample units $\mathcal{U}_t$ (e.g., schema--workload units for Text2SQL and dataset--class units for images). For each unit $u\in\mathcal{U}_t$, $n^{\text{gen}}_{u}$ denotes the number of generated candidates, $n^{\text{val}}_{u}$ denotes the number of candidates that are filtered/validated, and $n^{\text{exec}}_{u}$ denotes the number of SQL executions used for verification (with $n^{\text{exec}}_{u}{=}0$ for $\textsc{img}$). With per-operation unit costs $c_t^{\text{gen}}$, $c_t^{\text{val}}$, and $c_t^{\text{exec}}$, the total cost is computed as:
\begin{equation}
C=\sum_{t\in\{\textsc{sql},\textsc{img}\}}\sum_{u\in\mathcal{U}_t}\!\left(n^{\text{gen}}_{u}c_t^{\text{gen}}+n^{\text{val}}_{u}c_t^{\text{val}}+n^{\text{exec}}_{u}c_t^{\text{exec}}\right)\le B.
\label{eq:budget_metadataset}
\end{equation}
For Text2SQL ($3.4\text{M}$ samples), unit costs $\{c^{\text{gen}}, c^{\text{val}}, c^{\text{exec}}\}{=}\{8, 2, 40\}{\times}10^{-5}$ yield a projected total $C_{\textsc{sql}}{\approx}489.6$. For Images ($2.5\text{M}$ samples), costs $\{c^{\text{gen}}, c^{\text{val}}\}{=}\{15, 3\}{\times}10^{-5}$ yield $C_{\textsc{img}}{\approx}456.8$. Thus, the total estimated cost is $946.4{\le}B$. We enforce strict per-sample caps $\bar n_t^{\text{gen}}$ and $\bar n_t^{\text{exec}}$ (e.g., $\bar n_{\textsc{sql}}^{\text{gen}}{=}160$, $\bar n_{\textsc{sql}}^{\text{exec}}{=}40$, $\bar n_{\textsc{img}}^{\text{gen}}{=}300$) to guarantee the worst-case.

\sstitle{Training Data Formation}
From MetaDataset, we construct workloads for meta-learning by partitioning the corpus into $\mathcal{D}_{\mathrm{train}}$ and $\mathcal{D}_{\mathrm{val}}$ and repeatedly sampling target subsets $s_i$ from these splits. Each sampled subset $s_i$ captures a distinct distribution shift scenario relative to a model's original training data. For Text2SQL, $s_i$ varies database schemas, SQL operators, and linguistic forms. For Image Classification, it varies class subsets, background domains, and acquisition styles. We generate 30K such target subsets, each with $|s_i|{=}10$K. This balances coverage and computational cost while spanning both small-scale human-curated regimes and the large synthetic corpora, as summarized in \autoref{fig:tsne_all} and \autoref{tab:datasets_all}.

\subsection{Data Coverage}
\label{sec:experiment_data_coverage}

\begin{table}[t]
\centering
\small
\setlength{\tabcolsep}{6pt}
\resizebox{0.9\linewidth}{!}{%
\begin{tabular}{l l r}
\toprule
\multicolumn{3}{l}{\textbf{Text2SQL}} \\
\midrule
Dataset & Source & \# Examples \\
\midrule
WikiSQL~\cite{zhong2017seq2sql} & Human+Template & 80{,}654 \\
Spider~\cite{yu-etal-2018-spider} & Human & 10{,}181 \\
SParC~\cite{yu2019sparc} & Human & 12{,}726 \\
CoSQL~\cite{yu2019cosql} & Human & 10{,}000{+} \\
BIRD~\cite{li2023bird} & Human & 12{,}751 \\
ScienceBenchmark~\cite{stockinger2023sciencebenchmark} & Hybrid & 5{,}031 \\
EHRSQL~\cite{lee2022ehrsql} & Human+Template & 20{,}108 \\
KaggleDBQA~\cite{lee2021kaggledbqa} & Human & 272 \\
SynSQL-2.5M~\cite{li2025omnisql} & LLM-Gen & 2{,}544{,}390 \\
\textbf{MetaDataset (ours)} & Synthetic & 3{,}373{,}204 \\
\midrule
\multicolumn{3}{l}{\textbf{Image Classification}} \\
\midrule
MNIST~\cite{lecun2002mnist} & Handwritten & 70{,}000 \\
USPS~\cite{hull2002usps} & Real-world (scanned) & 9{,}298 \\
SVHN~\cite{netzer2011svhn} & Real-world (street-view) & 99{,}289 \\
COCO 2017~\cite{lin2014coco} & Human-annotated & 123{,}287 \\
PASCAL VOC 2012~\cite{everingham2010pascal} & Human-annotated & 11{,}540 \\
ImageNet ILSVRC12~\cite{deng2009imagenet} & Curated Web & 1{,}331{,}167 \\
\textbf{MetaDataset (ours)} & Synthetic & 2{,}487{,}936 \\
\bottomrule
\end{tabular}}
\caption{Dataset sizes and source categories used in the coverage analysis for Text2SQL and Image Classification.}
\label{tab:datasets_all}
\vspace{-1em}
\end{table}

\begin{figure}[t]
  \centering
  \begin{subfigure}{\linewidth}
    \centering
    \scalebox{1}[0.9]{\includegraphics[width=0.8\linewidth]{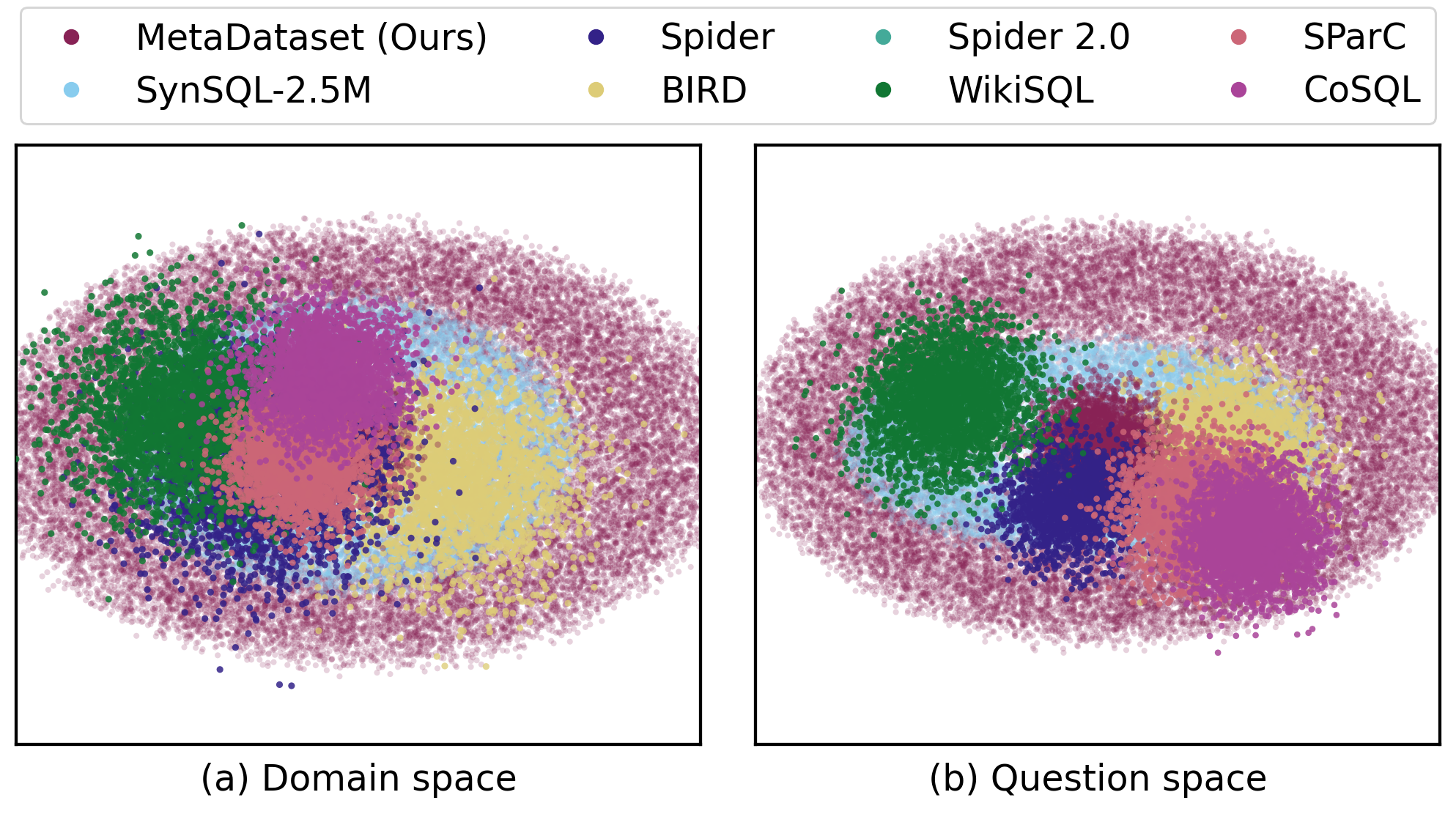}}
    \vspace{-1em}
    \caption{Text2SQL}
    \label{fig:tsne_text2sql}
  \end{subfigure}
  \begin{subfigure}{\linewidth}
    \centering
    \scalebox{1}[0.9]{\includegraphics[width=0.8\linewidth]{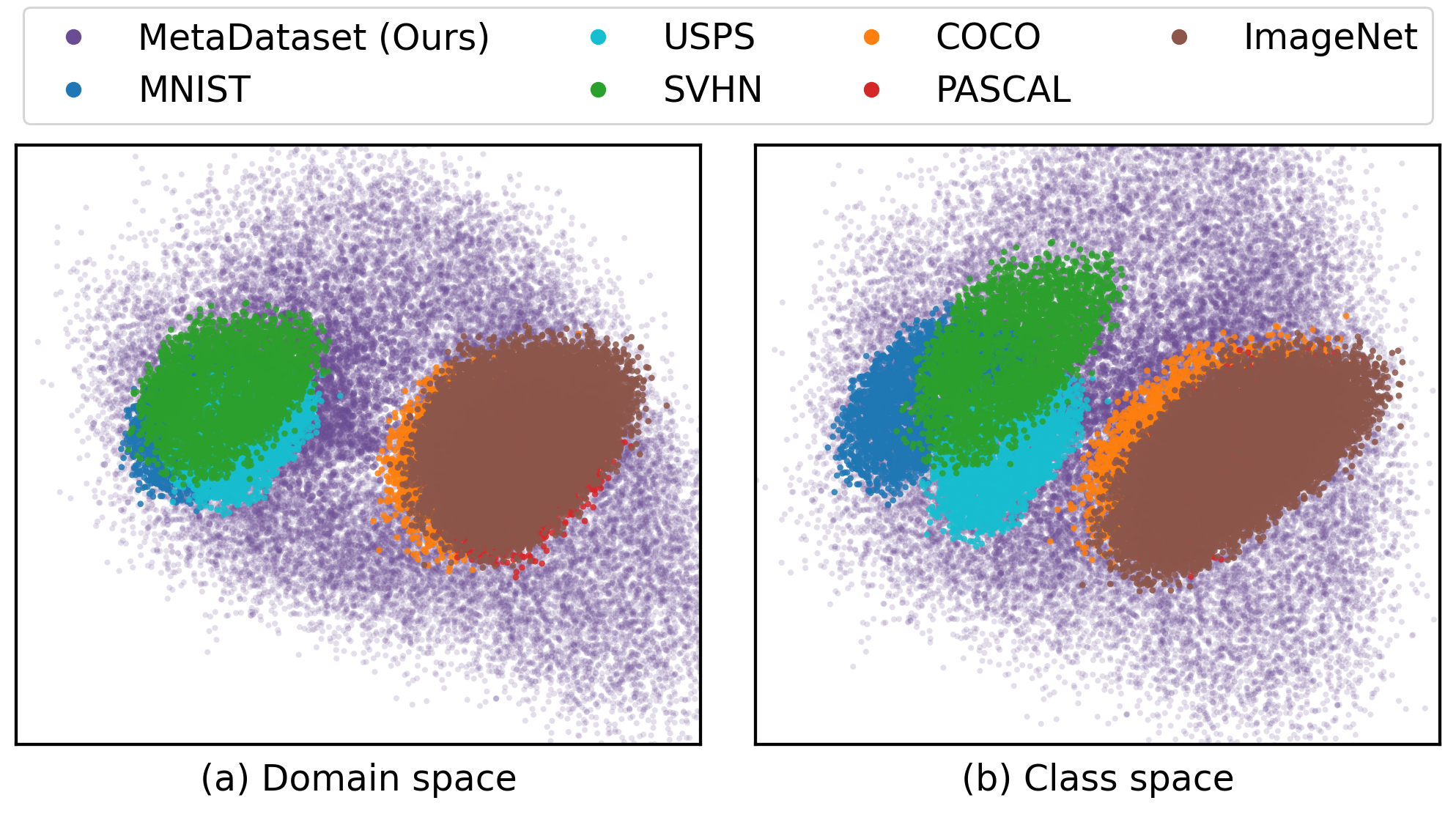}}
    \vspace{-1em}
    \caption{Image Classification}
    \label{fig:tsne_image}
  \end{subfigure}
  \caption{t-SNE of semantic coverage across modalities.}
  \label{fig:tsne_all}
  \vspace{-1em}
\end{figure}

To answer \textbf{RQ1 (Data Coverage)}, we assess whether \textit{MetaDataset} provides comprehensive and realistic coverage of deployment scenarios across Text2SQL and Image Classification, capturing diversity in schema structure, SQL composition, natural-language usage, as well as visual domains, styles, and semantic categories.

\sstitle{Data Size}
\autoref{tab:datasets_all} summarizes the scale of the datasets used in our coverage analysis across Text2SQL and Image Classification. Human-curated Text2SQL benchmarks are typically limited to at most tens of thousands of examples, with several datasets containing fewer than $15$K queries, whereas recent LLM-generated corpora such as SynSQL-2.5M~\cite{li2025omnisql} expand to millions of instances. Our MetaDataset further increases scale to over $3.3$M Text2SQL queries, exceeding all existing benchmarks. A similar pattern appears in Image Classification: classic handwritten and real-world digit datasets remain under $100$K examples, while large curated collections such as ImageNet ILSVRC12~\cite{deng2009imagenet} exceed one million images. MetaDataset reaches $2.49$M images, placing it among the largest resources used in this study and enabling systematic analysis of large-scale deployment regimes.

\sstitle{Semantic Coverage}
We visualize embedding-space geometry with t-SNE in \autoref{fig:tsne_all} to assess whether our construction spans realistic deployment scenarios across Text2SQL and Image Classification. In Text2SQL (\autoref{fig:tsne_text2sql}), WikiSQL~\cite{zhong2017seq2sql} forms a distinct single-table cluster, Spider~\cite{yu-etal-2018-spider}, SParC~\cite{yu2019sparc}, and CoSQL~\cite{yu2019cosql} group together under multi-table and conversational settings, and BIRD~\cite{li2023bird} occupies a separate region reflecting higher schema complexity, while Spider~2.0~\cite{lei2025spider} and SynSQL-2.5M~\cite{li2025omnisql} spread across multiple clusters, indicating broad schema and SQL coverage. Similar patterns appear in question space, where MetaDataset forms the widest envelope and fills gaps between benchmarks. For image classification (\autoref{fig:tsne_image}), digit datasets (MNIST~\cite{lecun2002mnist}, USPS~\cite{hull2002usps}, SVHN~\cite{netzer2011svhn}) separate from natural-image datasets (COCO~\cite{lin2014coco}, PASCAL~\cite{everingham2010pascal}, ImageNet~\cite{deng2009imagenet}) in background space, while class space reflects shared object semantics, and in both views MetaDataset overlaps all groups, capturing within-family and cross-family shifts. Overall, these visualizations qualitatively confirm broad semantic coverage across established benchmarks in both modalities.

\subsection{Evaluator Learning}
\label{sec:experiment_evaluator_learning}

\begin{table*}[t]
\centering
\small
\caption{MAE ($\downarrow$) of dataset-level accuracy estimation on unseen models across Text2SQL and Image Classification. Each cell reports mean $\pm$ 95\% CI (percentage points). Best in \textbf{bold}, second best \underline{underlined}.}
\label{tab:mae_unseen_multitask}
\setlength{\tabcolsep}{5pt}
\resizebox{0.8\textwidth}{!}{%
\begin{tabular}{l l cccccc}
\toprule
Tasks & Methods & Meta-Llama-3-70B & Qwen2.5-32B & XiYanSQL-14B & Ministral-3-14B & gemma-2-2b & Avg. \\
\midrule
\multirow{9}{*}{Text2SQL}
& DoC~\cite{guillory2021predicting}          & 15.42 $\pm$ 2.31 & 15.97 $\pm$ 2.45 & 16.18 $\pm$ 2.28 & 15.66 $\pm$ 2.39 & 16.05 $\pm$ 2.51 & 15.86 $\pm$ 2.39 \\
& ATC~\cite{garg2022leveraging}             & 17.21 $\pm$ 2.18 & 17.88 $\pm$ 2.34 & 18.06 $\pm$ 2.11 & 17.52 $\pm$ 2.27 & 17.95 $\pm$ 2.41 & 17.72 $\pm$ 2.26 \\
& AGD~\cite{jiang2022assessing}      & 14.77 $\pm$ 2.26 & 15.18 $\pm$ 2.39 & 14.96 $\pm$ 2.21 & 15.04 $\pm$ 2.30 & 15.25 $\pm$ 2.44 & 15.04 $\pm$ 2.32 \\
& PseudoAutoEval~\cite{boyeau2025autoeval}         & 13.84 $\pm$ 2.05 & 14.31 $\pm$ 2.22 & 14.58 $\pm$ 2.09 & 14.07 $\pm$ 2.16 & 14.42 $\pm$ 2.28 & 14.24 $\pm$ 2.16 \\
& AutoEval~\cite{deng2021labels}               & 11.62 $\pm$ 1.94 & 12.05 $\pm$ 2.08 & 12.33 $\pm$ 1.91 & 11.89 $\pm$ 2.01 & 12.21 $\pm$ 2.13 & 12.02 $\pm$ 2.01 \\
& NL2SQL-BUGS~\cite{Liu2025nl2sqlbugs}                    & \underline{9.31 $\pm$ 1.42} & \underline{9.68 $\pm$ 1.55} & \underline{9.84 $\pm$ 1.37} & \underline{9.47 $\pm$ 1.49} & \underline{9.76 $\pm$ 1.61} & \underline{9.61 $\pm$ 1.49} \\
& KNN                    & 9.94 $\pm$ 1.51 & 10.31 $\pm$ 1.64 & 10.12 $\pm$ 1.46 & 10.23 $\pm$ 1.58 & 10.30 $\pm$ 1.70 & 10.18 $\pm$ 1.58 \\
& Top-$k$                    & 9.51 $\pm$ 1.44 & 9.88 $\pm$ 1.57 & 9.67 $\pm$ 1.39 & 9.79 $\pm$ 1.51 & 9.85 $\pm$ 1.63 & 9.74 $\pm$ 1.51 \\
& \textbf{MetaEvaluator (Ours)}                      & \textbf{3.41 $\pm$ 0.78} & \textbf{3.76 $\pm$ 0.84} & \textbf{3.55 $\pm$ 0.71} & \textbf{3.69 $\pm$ 0.80} & \textbf{3.88 $\pm$ 0.89} & \textbf{3.66 $\pm$ 0.80} \\
\midrule
& 
& ResNeXt-50-32x4d & RegNetY-8GF & ConvNeXt-Tiny & ViT-Tiny & DeiT-Small & Avg. \\
\midrule
\multirow{9}{*}{Image Classification}
& DoC~\cite{guillory2021predicting}            & 16.03 $\pm$ 2.21 & 16.54 $\pm$ 2.37 & 16.27 $\pm$ 2.18 & 16.88 $\pm$ 2.46 & 17.02 $\pm$ 2.55 & 16.55 $\pm$ 2.35 \\
& ATC~\cite{garg2022leveraging}               & 17.02 $\pm$ 2.34 & 17.46 $\pm$ 2.51 & 17.19 $\pm$ 2.28 & 17.83 $\pm$ 2.59 & 17.96 $\pm$ 2.67 & 17.49 $\pm$ 2.48 \\
& AGD~\cite{jiang2022assessing}      & 15.11 $\pm$ 2.08 & 15.59 $\pm$ 2.22 & 15.32 $\pm$ 2.05 & 15.74 $\pm$ 2.31 & 15.97 $\pm$ 2.40 & 15.55 $\pm$ 2.21 \\
& PseudoAutoEval~\cite{boyeau2025autoeval}           & 13.67 $\pm$ 2.01 & 14.02 $\pm$ 2.15 & 13.81 $\pm$ 1.98 & 14.19 $\pm$ 2.24 & 14.37 $\pm$ 2.33 & 14.01 $\pm$ 2.14 \\
& AutoEval~\cite{deng2021labels}                 & 11.44 $\pm$ 1.86 & 11.79 $\pm$ 1.97 & 11.58 $\pm$ 1.82 & 12.01 $\pm$ 2.05 & 12.18 $\pm$ 2.12 & 11.80 $\pm$ 1.96 \\
& SelfTrainEns~\cite{chen2021detecting}                      & 10.02 $\pm$ 1.31 & 9.28 $\pm$ 1.43 & 10.11 $\pm$ 1.26 & 10.46 $\pm$ 1.51 & 10.63 $\pm$ 1.60 & 11.30 $\pm$ 1.42 \\
& KNN                    & 9.96 $\pm$ 1.42 & 10.28 $\pm$ 1.55 & 10.07 $\pm$ 1.37 & 10.31 $\pm$ 1.60 & 10.28 $\pm$ 1.66 & 10.18 $\pm$ 1.52 \\
& Top-$k$                    & \underline{9.55 $\pm$ 1.36} & \underline{9.83 $\pm$ 1.48} & \underline{9.66 $\pm$ 1.31} & \underline{9.82 $\pm$ 1.54} & \underline{9.84 $\pm$ 1.59} & \underline{9.74 $\pm$ 1.46} \\
& \textbf{MetaEvaluator (Ours)}                        & \textbf{3.58 $\pm$ 0.73} & \textbf{3.74 $\pm$ 0.81} & \textbf{3.61 $\pm$ 0.69} & \textbf{3.89 $\pm$ 0.88} & \textbf{3.97 $\pm$ 0.94} & \textbf{3.76 $\pm$ 0.81} \\
\bottomrule
\end{tabular}}
\vspace{-1em}
\end{table*}

Using MetaDataset from \autoref{sec:experiment_data_coverage}, we perform meta-learning and evaluate MetaEvaluator for label-free accuracy estimation, addressing \textbf{RQ2 (Evaluator Learning)}. The reference model pool is specified in our codebase, and none of these models overlap with the unseen models evaluated in subsequent experiments.

\sstitle{Evaluator Benchmark}
Results in \autoref{tab:mae_unseen_multitask} average over unseen source--target transfers in Text2SQL and Image Classification. Additional results on a broader set of models are provided in our public repository at \url{https://github.com/phkhanhtrinh23/MetaEvaluator}. MetaEvaluator attains the lowest MAE ($\approx3$--$4$), clearly outperforming all baselines; NL2SQL-BUGS is the strongest Text2SQL baseline but struggles on harder targets such as Spider~2.0 and BIRD, while SelfTrainEns is the strongest vision baseline yet remains less accurate and more costly. We also compare two retrieval baselines built on the same shift descriptors: KNN averages the performances of the unseen model on the $k$ nearest labeled datasets, and Top-$k$ averages the top-$k$ most similar labeled datasets using cosine similarity. Both remain far behind MetaEvaluator, showing that the gains come from learning a transferable evaluation function rather than simply retrieving nearby labeled workloads. The same result holds for five black-box models accessed through text outputs only, where estimated shift descriptors are computed with a frozen BERT encoder and all baselines receive the same descriptor access for fairness. In this setting, MetaEvaluator remains the most accurate, with compared to LTV and DoC, which indicates SDs can capture meaningful distribution shifts even with model outputs only.

\begin{table}[t]
\centering
\small
\caption{Impact of meta-learning algorithm on evaluation accuracy and cost. Each MAE cell reports mean $\pm$ 95\% CI; best is in \textbf{bold} and second best is \underline{underlined}.}
\label{tab:ablation_meta_alg}
\resizebox{0.8\linewidth}{!}{%
\begin{tabular}{l c c c}
\toprule
& MAE & \# Steps & \# Extra params (M) \\
\midrule
MAML~\cite{finn2017maml} & 11.63 $\pm$ 1.21 & 12 & 0.00 \\
FO-MAML~\cite{finn2017maml} & 8.88 $\pm$ 1.29 & 10 & 0.00 \\
Reptile~\cite{nichol2018reptile} & 9.12 $\pm$ 1.34 & 9 & 0.00 \\
Meta-SGD~\cite{li2017metasgd} & \underline{8.21 $\pm$ 1.18} & 9 & 0.38 \\
ANIL~\cite{raghu2020anil} & 8.47 $\pm$ 1.25 & 8 & 0.00 \\
% CAVIA~\cite{zintgraf2019fast} & 8.62 $\pm$ 1.22 & 6 & 0.10 \\
ProtoNet~\cite{snell2017prototypical} & 12.45 $\pm$ 1.41 & 0 & 0.00 \\
MetaEvaluator (Ours) & \textbf{3.26 $\pm$ 0.96} & 3 & 0.12 \\
\bottomrule
\end{tabular}}
\vspace{-1em}
\end{table}

\begin{figure}[t]
    \centering
    \includegraphics[width=0.9\linewidth]{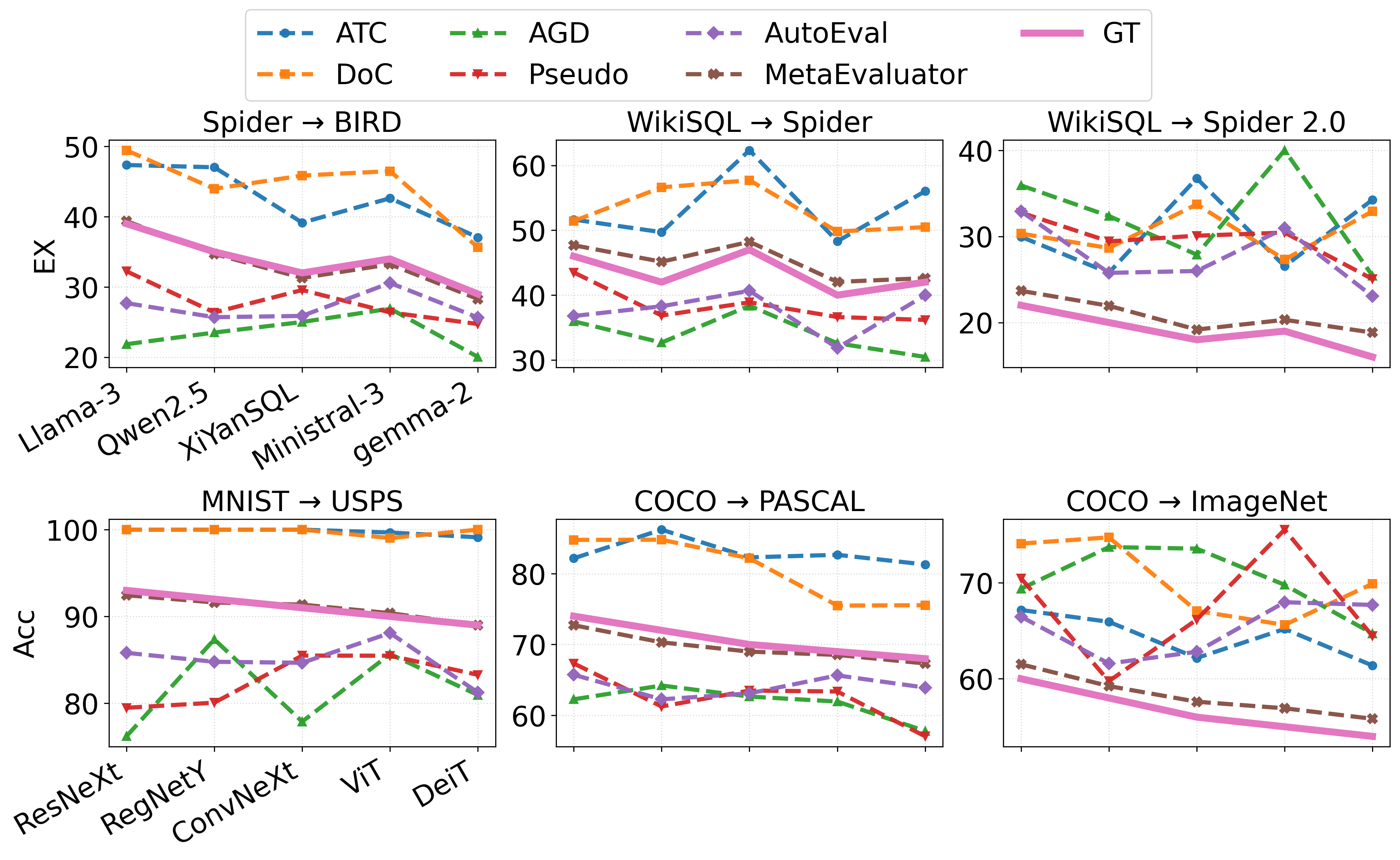}
    \caption{Calibration of accuracy estimation across transfers.}
    \label{fig:calibration_multitask}
    \vspace{-1em}
\end{figure}

\begin{figure}[t]
    \centering
    \includegraphics[width=0.8\linewidth]{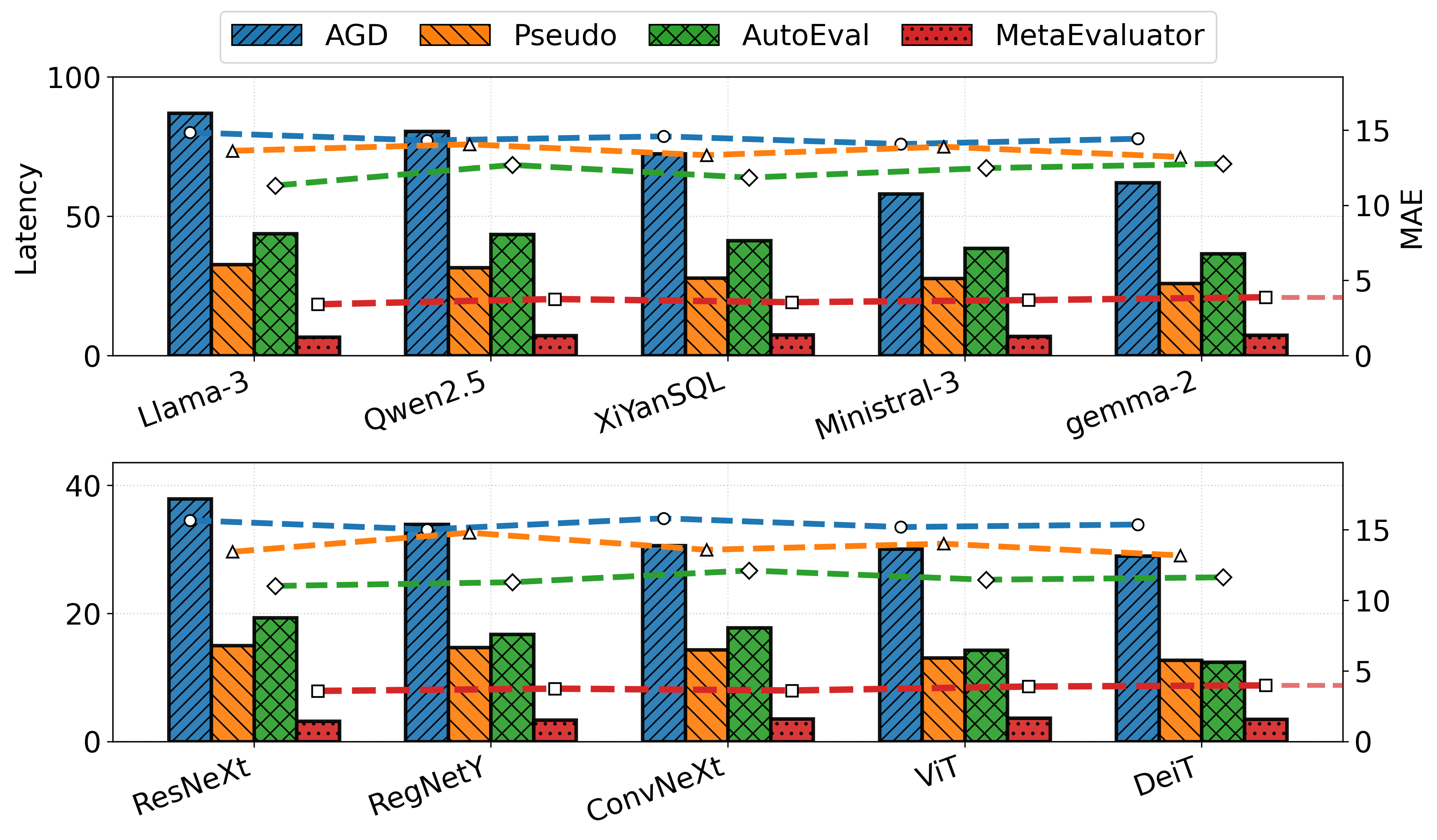}
    \caption{Latency--MAE trade-offs on unseen models.}
    \label{fig:latency_mae_multitask}
    \vspace{-1em}
\end{figure}

\begin{figure}[t]
    \centering
    \includegraphics[width=0.8\linewidth]{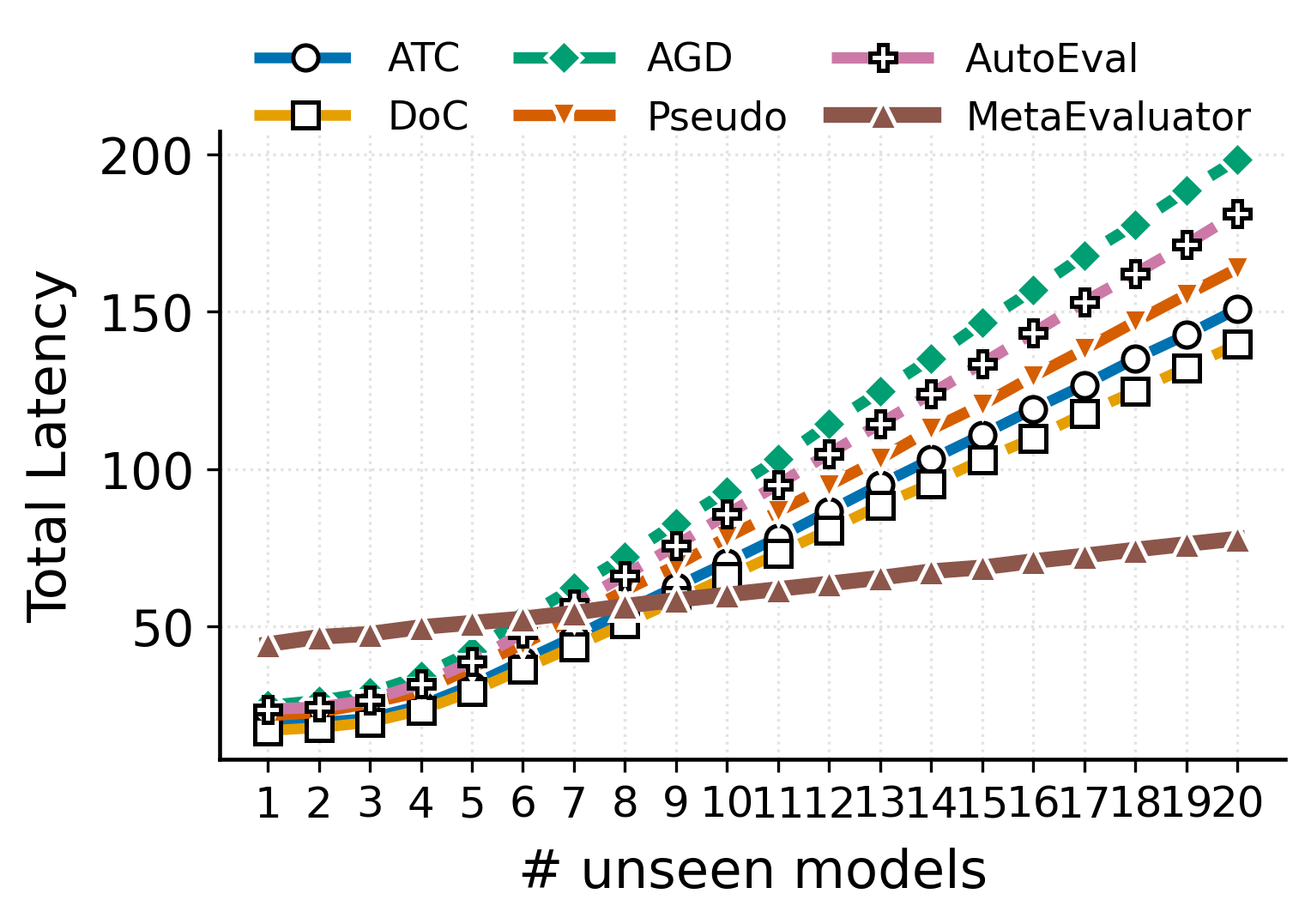}
    \caption{Total training and evaluation latency as the number of unseen models increases.}
    \label{fig:total_latency}
    \vspace{-1em}
\end{figure}

\begin{figure}[t]
    \centering
    \includegraphics[width=0.8\linewidth]{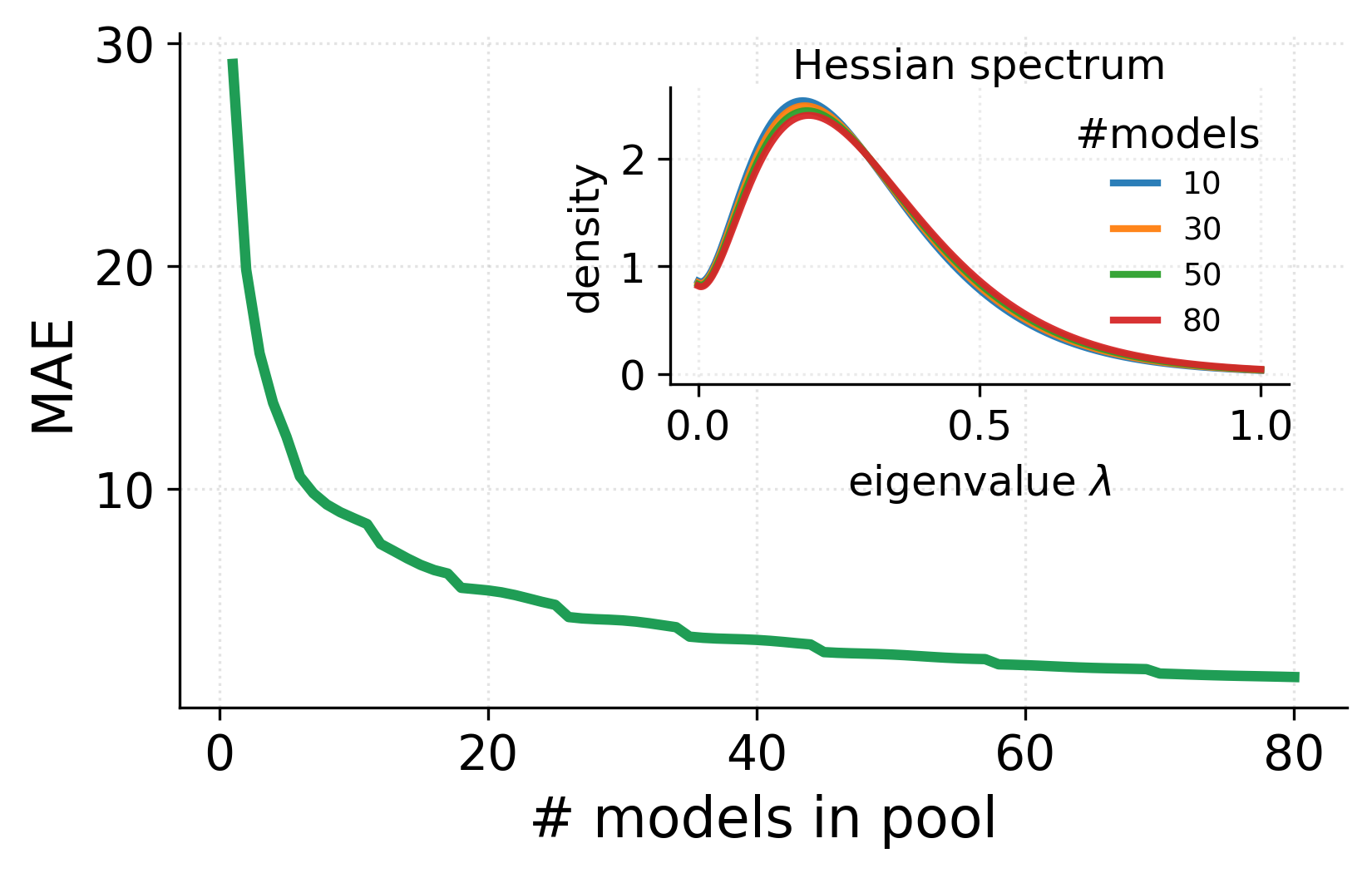}
    \caption{Meta-learning improves with pool size. Inset: Hessian spectra remain stable as pool size increases.}
    \label{fig:mae_vs_pool_size_with_hessian_inset}
    \vspace{-1em}
\end{figure}

\begin{figure}[t]
    \centering
    \includegraphics[width=0.8\linewidth]{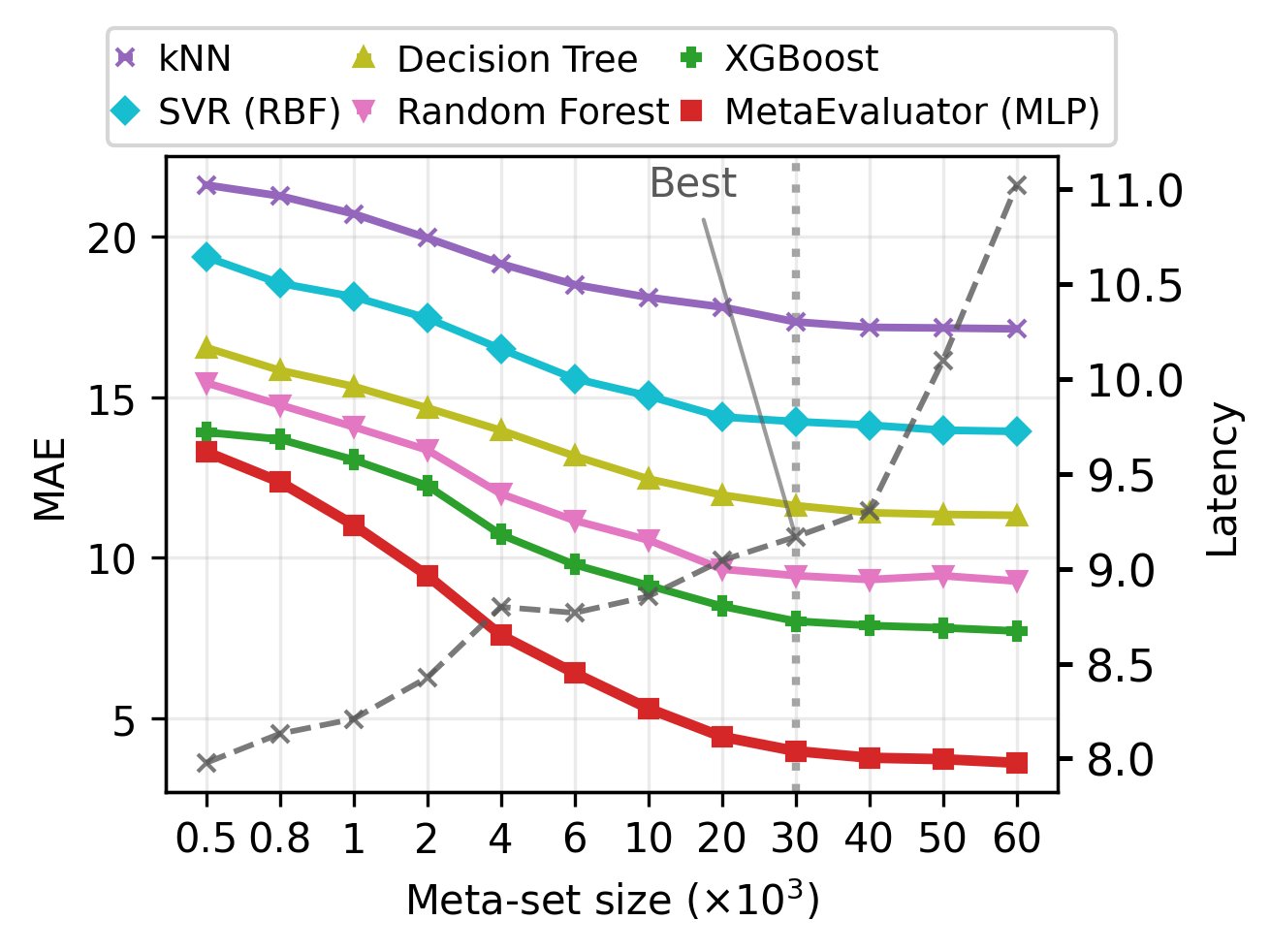}
    \caption{MLP attains the lowest error and benefits most from larger meta-sets, while costs rise sharply beyond $30$K with marginal gains.}
    \label{fig:meta_set_size}
    \vspace{-1em}
\end{figure}

\begin{figure}[t]
    \centering
    \includegraphics[width=0.8\linewidth]{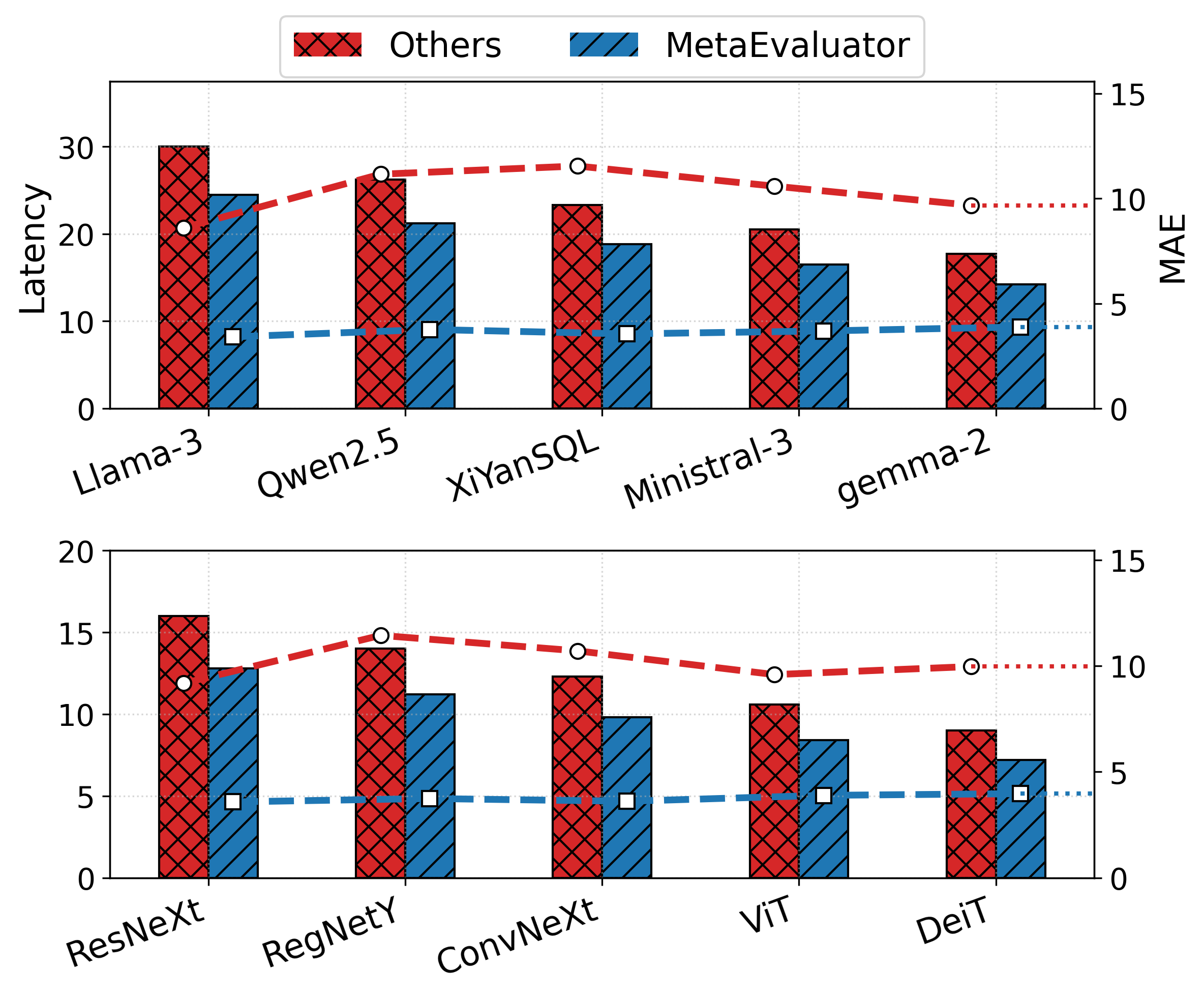}
    \caption{MetaEvaluator consistently reduces both MAE and latency compared to other meta-learning algorithms.}
    \label{fig:meta_learning_algorithms}
    \vspace{-1em}
\end{figure}

\sstitle{Estimation Calibration}
\autoref{fig:calibration_multitask} compares predictions with ground truth across transfers. MetaEvaluator tracks GT most closely on both tasks, with small deviations on easier shifts and controlled bias on harder ones, whereas ATC and DoC systematically overestimate and other baselines fluctuate widely. This stable calibration explains the low MAE observed in \autoref{tab:mae_unseen_multitask}.

\begin{table}[t]
\centering
\small
\setlength{\tabcolsep}{4pt}
\caption{MAE ($\downarrow$) for black-box large models accessed through text outputs only. Each cell reports mean $\pm$ 95\% CI (percentage points). Best in \textbf{bold}, second best \underline{underlined}.}
\label{tab:blackbox_mae}
\resizebox{\linewidth}{!}{%
\begin{tabular}{l cccccc}
\toprule
Methods & GPT-5.1 & Claude Opus 4.5 & Gemini 3 Pro & Qwen2.5-72B & Mistral-7B & Avg. \\
\midrule
DoC~\cite{guillory2021predicting}     & 16.55 $\pm$ 2.39 & 17.12 $\pm$ 2.51 & 16.71 $\pm$ 2.34 & 16.93 $\pm$ 2.46 & 16.89 $\pm$ 2.45 & 16.84 $\pm$ 2.43 \\
ATC~\cite{garg2022leveraging}         & 17.63 $\pm$ 2.57 & 18.21 $\pm$ 2.69 & 17.79 $\pm$ 2.52 & 18.02 $\pm$ 2.64 & 17.95 $\pm$ 2.63 & 17.92 $\pm$ 2.61 \\
AGD~\cite{jiang2022assessing}         & 14.96 $\pm$ 2.34 & 15.51 $\pm$ 2.46 & 15.12 $\pm$ 2.29 & 15.34 $\pm$ 2.41 & 15.22 $\pm$ 2.40 & 15.23 $\pm$ 2.38 \\
BBSE~\cite{lipton2018bbse}            & 13.11 $\pm$ 3.13 & 13.68 $\pm$ 3.26 & 13.29 $\pm$ 3.08 & 13.54 $\pm$ 3.21 & 13.48 $\pm$ 3.22 & 13.42 $\pm$ 3.18 \\
LTV~\cite{renggli2019ltv}             & \underline{10.58 $\pm$ 2.59} & \underline{11.13 $\pm$ 2.71} & \underline{10.74 $\pm$ 2.55} & \underline{10.96 $\pm$ 2.66} & \underline{10.94 $\pm$ 2.69} & \underline{10.87 $\pm$ 2.64} \\
\textbf{MetaEvaluator (Ours)}         & \textbf{4.52 $\pm$ 0.97} & \textbf{4.88 $\pm$ 1.07} & \textbf{4.61 $\pm$ 0.93} & \textbf{4.79 $\pm$ 1.03} & \textbf{4.75 $\pm$ 1.05} & \textbf{4.71 $\pm$ 1.02} \\
\bottomrule
\end{tabular}}
\vspace{-1em}
\end{table}

\subsection{Benchmarking Capability}
\label{sec:experiment_benchmarking_capability}

To answer \textbf{RQ3 (Benchmarking Capability)}, we evaluate whether MetaEvaluator supports fast and lightweight benchmarking, and how it scales as the reference model pool continues to expand.

\sstitle{Evaluation Latency}
The best 3 methods from \autoref{tab:mae_unseen_multitask} are selected to compare the evaluation latency. \autoref{fig:latency_mae_multitask} shows that Text2SQL (\autoref{fig:latency_mae_multitask}) incurs substantially higher cost than Image Classification due to LLM decoding and schema-conditioned reasoning. Training-based baselines (AGD, PseudoAutoEval, AutoEval) remain slow because each new model triggers retraining. In contrast, MetaEvaluator stays both the fastest (about 1--2 minutes per model) and the most accurate. Moreover, \autoref{fig:total_latency} confirms its much flatter growth as the number of unseen models increases. By amortizing training latency across the reference pool and requiring only a few lightweight adaptation steps and forward passes per model, MetaEvaluator enables practitioners to benchmark large streams of candidate Image Classification or Text2SQL systems quickly on unlabeled data, placing it on a strictly better accuracy--latency Pareto frontier.

\begin{table}[t]
\centering
\small
\setlength{\tabcolsep}{5pt}
\caption{Per-component cost decomposition. One-time costs (data construction, meta-training) are amortized across all future models.}
\label{tab:deployment_cost}
\resizebox{0.95\linewidth}{!}{%
\begin{tabular}{l c c c}
\toprule
Component & AGD & PseudoAutoEval & MetaEvaluator \\
\midrule
Data construction (once)   & N/A         & N/A         & $\sim$2 hrs \\
Meta-training (once)       & N/A         & N/A         & $\sim$1 hr \\
\midrule
Inference (per model)      & $\sim$1.2 min & $\sim$1.0 min & $\sim$0.3 min \\
SD computation (per model) & N/A         & N/A         & $\sim$0.4 min \\
Retraining/adaptation (per model) & $\sim$3.2 hrs & $\sim$2.8 hrs & $\sim$0.9 min \\
\midrule
\textbf{Total per new model} & $\sim$3.2 hrs & $\sim$2.8 hrs & \textbf{$\sim$1--2 min} \\
\bottomrule
\end{tabular}}
\vspace{-1em}
\end{table}

\sstitle{Cost Decomposition}
To clarify the source of MetaEvaluator's efficiency, \autoref{tab:deployment_cost} decomposes the end-to-end cost into one-time setup and per-model deployment cost. MetaDataset construction ($\sim$2 hrs) and meta-training ($\sim$1 hr) are incurred only once and amortized across every future model. At deployment, MetaEvaluator requires only a lightweight forward pass, shift-descriptor computation, and a few context-adaptation steps, totaling $\sim$1--2 minutes per new model. In contrast, training-based baselines retrain or run auxiliary inference for each arriving model, incurring slowdowns of several hours per model.

\subsection{Ablation Study}
\label{sec:experiment_ablation}

To answer \textbf{RQ4 (Component Utility)}, we examine how MetaEvaluator's design choices support fast and lightweight benchmarking.

\sstitle{Meta-set Size}
% We study how regression architecture and meta-set size $N$ trade off estimation accuracy and training cost in realistic deployment scenarios.
Each meta instance $(\mathcal{D}_{\mathrm{train}},s_i)$ encodes a train--test shift through $\mathrm{SD}_{\mathrm{train}}=h(\phi_{\mathcal{D}_{\mathrm{train}}},\phi_{s_i})$ (\autoref{eq:shift_descriptors}). As shown in \autoref{fig:meta_set_size}, we compare MetaEvaluator's MLP with classical regressors while varying $N$. Estimation error decreases for all methods as $N$ grows, but the MLP continues improving up to $N{=}30$K while simpler models saturate earlier, and training cost rises sharply beyond this point with only marginal accuracy gains. This enables practitioners to select $N$ to meet target MAE levels while keeping training cost compatible with rapid deployment needs.

\sstitle{Meta-learning Algorithms}
We compare meta-learning algorithms under the same MetaEvaluator setup: the three-layer MLP backbone, shift descriptors, MetaDataset splits, and training budget are fixed, so methods differ only in inner-loop adaptation. MAML~\cite{finn2017maml} and FO-MAML~\cite{finn2017maml} adapt the full backbone $\theta$ using second- and first-order meta-gradients, requiring $12$ and $10$ steps, while Reptile~\cite{nichol2018reptile} uses the inner-loop SGD trajectory in $9$ steps. Meta-SGD~\cite{li2017metasgd} additionally learns per-parameter step sizes, adding $0.38$M parameters. In contrast, MetaEvaluator adapts only a $512$-dimensional context vector ($0.12$M extra parameters), achieving the lowest MAE in just $3$ steps. \autoref{fig:meta_learning_algorithms} confirms that it also yields the best MAE--latency trade-off across Text2SQL and Image Classification, supporting fast and reliable benchmarking of unseen models on unlabeled workloads.

\begin{figure}[t]
  \centering
  \includegraphics[width=0.49\linewidth]{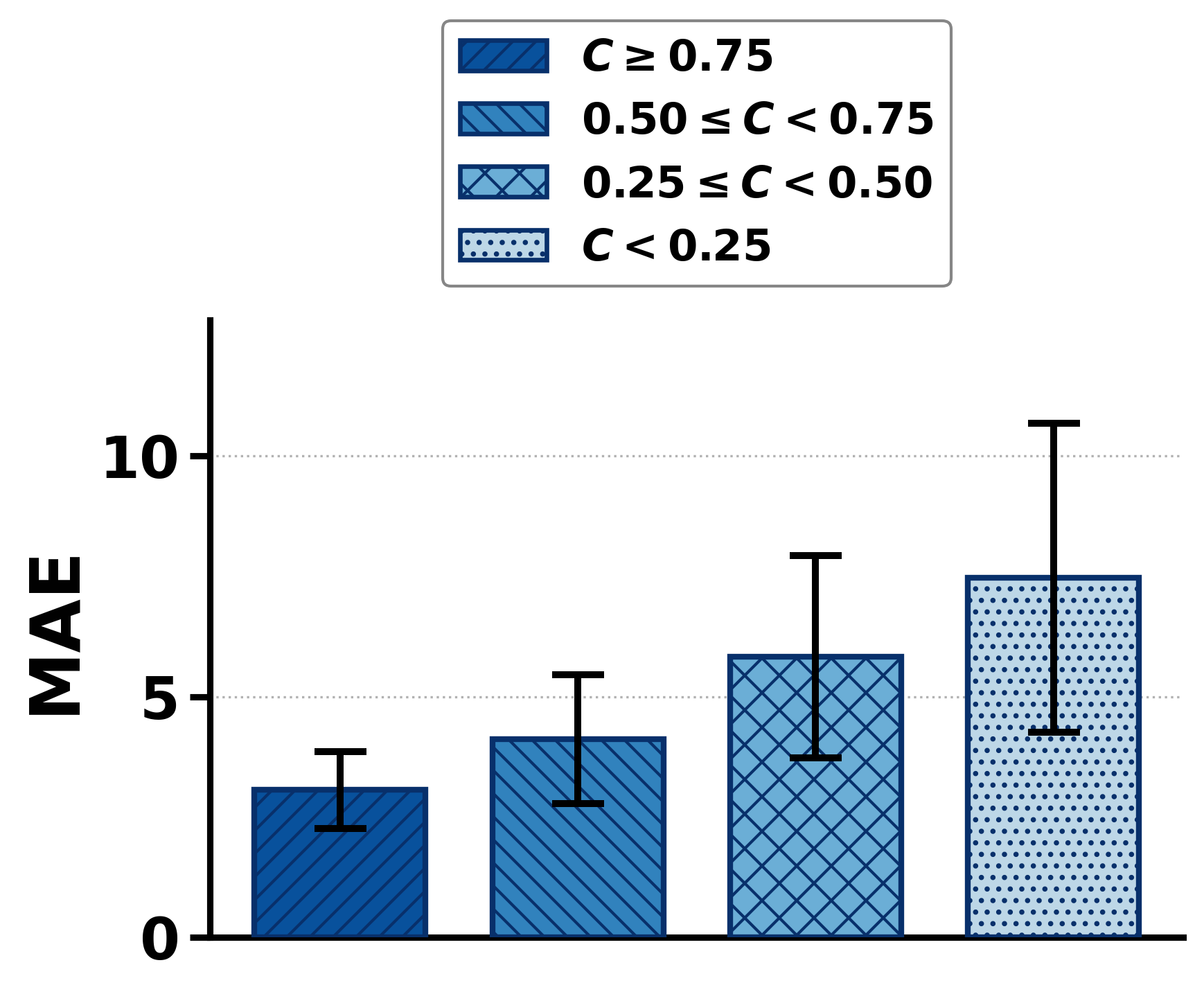}
  \hfill
  \includegraphics[width=0.49\linewidth]{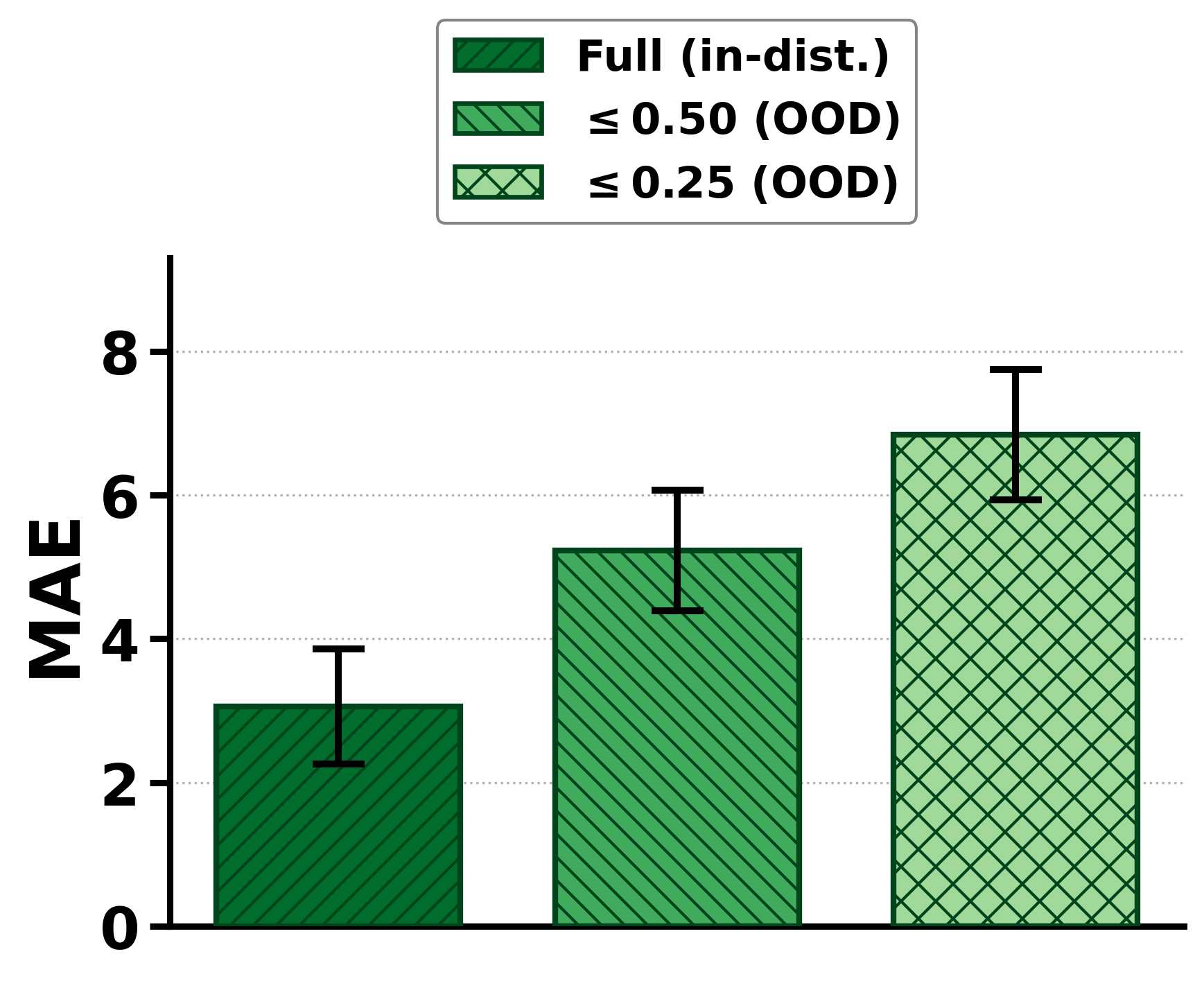}
  \caption{Robustness of MetaEvaluator under varying reference-pool coverage (left) and OOD evaluation (right). Black error bars indicate 95\% CIs.}
  \label{fig:robustness}
  \vspace{-1em}
\end{figure}

\sstitle{Coverage Robustness}
\autoref{fig:robustness} shows graceful degradation under both lower reference-pool coverage and stricter OOD evaluation. MAE rises from $3.06$ to $7.47$ as the coverage ratio $C=|\mathcal{M}_{\mathrm{train}}\cap\mathcal{A}|/|\mathcal{A}|$ drops from $C\!\ge\!0.75$ to $C\!<\!0.25$, where $\mathcal{M}_{\mathrm{train}}$ is the reference model pool used for meta-training and $\mathcal{A}$ is the set of unseen test architectures, and from $3.06$ to $5.23$ and $6.84$ when MetaDataset coverage is reduced to $\le0.50$ and $\le0.25$. Following the standard MAML episodic setting, adaptation for every architecture variant uses the same labeled split $\mathcal{D}_{\mathrm{train}}$. Changing the split across variants would misalign the $\theta^{\star}$ (\autoref{alg:meta_learning_algorithm}) and move the adapted context away from the learned context subspace. Real deployments rarely provide dense coverage over all future architectures or workload shifts. Even in these harder regimes, MetaEvaluator's MAE remains below the second-best baselines in \autoref{tab:mae_unseen_multitask}, indicating that it learns transferable evaluation behavior rather than memorizing the training pool.

\section{Conclusion and Future Work}
\label{sec:conclusion}
% \textcolor{blue}{Continue}

We introduced MetaEvaluator, a label-free framework for estimating the performance of unseen models on unseen and unlabeled workloads across Text2SQL and Image Classification. To our knowledge, MetaEvaluator is the first framework to explicitly address this double challenge: estimating dataset-level performance for newly released models on target workloads where labels are unavailable. By combining meta-learning with compact shift descriptors, MetaEvaluator amortizes evaluation knowledge across reference models and substantially reduces both estimation error and evaluation cost. Extensive experiments show that MetaEvaluator achieves a strong accuracy--efficiency trade-off, generalizes to held-out models and workloads, and scales as the pool of reference and candidate models grows. These results position MetaEvaluator as a practical and scalable tool for routine model screening in rapidly evolving model ecosystems, especially when target labeling is constrained by cost, privacy, or distribution drift. Future work will extend the framework to broader modalities and task families, further reduce dependence on labeled reference data, and integrate MetaEvaluator into multi-agent pipelines where low-cost evaluation can guide data selection, adaptation, debugging, and deployment decisions.

% We introduced MetaEvaluator, a label-free method for estimating the performance of unseen models on unseen workloads across Text2SQL and Image Classification. By combining meta-learning with optimal-transport alignment and compact shift descriptors, MetaEvaluator amortizes evaluation cost across reference models and achieves substantially lower error and latency than prior confidence-based and hypernetwork approaches. Extensive experiments show that MetaEvaluator consistently dominates the accuracy--efficiency trade-off and scales as reference and new models grow. These results position MetaEvaluator as a practical tool for continuous deployment selection of rapidly evolving models, without requiring additional annotation or per-model retraining.

% In future, MetaEvaluator can include additional modalities and task families beyond those studied here, and explore joint evaluation settings in which multiple candidate models are assessed simultaneously to further reduce deployment overhead in large model pools.

\section*{Acknowledgement}
The Australian Research Council partially supports this work under the streams of the Discovery Project (Grant No. DP240101108 and DP260100326), and the Linkage Project (Grant No. LP240200546).

% \begin{acks}
% \end{acks}

\bibliographystyle{ACM-Reference-Format}
\bibliography{ref}

\appendix
% \input{appendix_ot}
% \input{appendix_sql}
% \input{appendix_image}
% \input{appendix_c}









\end{document}